# Extensions of Simple Conceptual Graphs:
# the Complexity of Rules and Constraints

**Jean-François Baget**                                    BAGET@LIRMM.FR
**Marie-Laure Mugnier**                                 MUGNIER@LIRMM.FR
*LIRMM (CNRS & UM II), 161 rue Ada*
*34392 Montpellier, Cédex 5, France*

## Abstract

*Simple conceptual graphs* are considered as the kernel of most knowledge representation formalisms built upon Sowa's model. Reasoning in this model can be expressed by a graph homomorphism called projection, whose semantics is usually given in terms of positive, conjunctive, existential FOL. We present here a family of extensions of this model, based on *rules* and *constraints*, keeping graph homomorphism as the basic operation. We focus on the formal definitions of the different models obtained, including their operational semantics and relationships with FOL, and we analyze the decidability and complexity of the associated problems (consistency and deduction). As soon as rules are involved in reasonings, these problems are not decidable, but we exhibit a condition under which they fall in the polynomial hierarchy. These results extend and complete the ones already published by the authors. Moreover we systematically study the complexity of some particular cases obtained by restricting the form of constraints and/or rules.

## 1. Introduction

Conceptual graphs (CGs) have been proposed as a knowledge representation and reasoning model, mathematically founded both on logics and graph theory (Sowa, 1984). Though they have been mainly studied as a graphical interface for logics or as a diagrammatic system of logics (for instance, see Wermelinger, 1995, for general CGs equivalent to FOL), their graph-theoretic foundations have been less investigated. Most works in this area are limited to *simple conceptual graphs*, or *simple graphs* (Sowa, 1984; Chein & Mugnier, 1992), which correspond to the positive, conjunctive and existential fragment of FOL without functions. This model has three fundamental characteristics:

1. objects are bipartite *labelled graphs* (nodes represent *entities* and *relations* between these entities);

2. reasonings are based on graph-theoretic operations, relying on a kind of graph homomorphism called *projection*;

3. it is logically founded, reasonings being sound and complete w.r.t. FOL semantics, usually by way of the translation called Φ.

Main extensions of the simple graphs model, keeping graph homomorphism based operations and sound and complete semantics, are *inference rules* (Gosh & Wuwongse, 1995; Salvat & Mugnier, 1996; Salvat, 1998) and *nested graphs* (Chein, Mugnier, & Simonet, 1998; Preller, Mugnier, & Chein, 1998); for *general CGs* equivalent to FOL, an original deduction system (Kerdiles, 1997) combines analytic tableaux with the simple graphs projection. Some kind of *constraints* have been proposed to validate a knowledge base composed of simple graphs (Mineau & Missaoui, 1997; Dibie, Haemmerlé, & Loiseau, 1998).

We present here a family of extensions of the simple graphs model. The common ground for these extensions is that objects are *colored simple graphs* representing *facts*, *rules* or *constraints*, and operations are based upon projection. Given a knowledge base $\mathcal{K}$ and a simple graph $Q$ (which may represent a query, a goal, ..., depending on the application), the *deduction problem* asks whether $Q$ can be deduced from $\mathcal{K}$. According to the kinds of objects considered in $\mathcal{K}$, different reasoning models are obtained, composing the $\mathcal{SG}$ family. Though similar notions of rules and constraints can be found in the CG literature, their combination in reasonings had never been studied. One interest of our approach thus resides in providing a unifying framework combining rules and constraints in different ways.

In this paper, we focus on the formal definitions of these models, including their operational semantics and relationships with FOL, and we study the decidability and complexity of their associated decision problems, namely consistency and deduction. These results extend and complete the ones already published by the authors (Baget & Mugnier, 2001). Though both consistency and deduction are undecidable in the most general model of this family, we had already used a decidable subset of rules to solve the SISYPHUS-I problem, a test-bed proposed in the knowledge acquisition community (Baget, Genest, & Mugnier, 1999). We present here for the first time a detailed analysis of complexity when we restrict the knowledge base to this kind of rules (called *range restricted* rules). We also study particular cases of constraints.

In section 2 basic definitions and results about simple graphs are recalled. Section 3 presents an overview of the $\mathcal{SG}$ family. In particular, we explain why we consider graphical features of the simple graphs model as essential for knowledge modeling and point out that these properties are preserved in the $\mathcal{SG}$ family. In next sections we study the different members of the family. Rules are introduced in section 4, constraints in section 5, and section 6 studies models combining rules and constraints. As soon as rules are involved in reasonings, the associated decision problems are not decidable, but we exhibit a condition (finite expansion sets) under which computations always stop. In the particular case of *range restricted* rules, the complexity of these problems fall into the polynomial hierarchy. Section 7 is devoted to these decidable cases. In section 8, relationships with other works are established. In particular we point out algorithmic connections with constraint satisfaction problems (CSP) and show that the problem of checking the consistency of a knowledge base composed of simple graphs and constraints ($\mathcal{SGC}$-consistency) is equivalent to that of deciding the consistency of a mixed CSP (MIXED-SAT, Fargier, Lang, & Schiex, 1996).

## 2. Basic Notions: the $\mathcal{SG}$ Model

We recall in this section basic notions about simple conceptual graphs (Sowa, 1984; Chein & Mugnier, 1992). These graphs are considered as the kernel for most knowledge representation formalisms built upon Sowa's work. They are also the basic model for the $\mathcal{SG}$ family.

## 2.1 Definitions and Notations

Basic ontological knowledge is encoded in a structure called a *support*. Factual knowledge is encoded into *simple graphs* (SGs), defined with respect to a given support. A SG is a bipartite labelled graph (strictly speaking, it is a *multigraph*, since there can be several edges between two nodes). One class of nodes represents entities, the other class represents relationships between these entities. Nodes are labelled by elements of the support. Elementary reasonings are computed by a graph homomorphism called *projection*.

**Definition 1 (Support)** *A support is a 4-tuple* $\mathcal{S} = (T_C, T_R, \mathcal{I}, \tau)$. $T_C$ *and* $T_R$ *are two partially ordered finite sets, respectively of* concept types *and* relation types. $T_R$ *is partitioned into subsets* $T_R^1 \dots T_R^k$ *of relation types of arity* $1 \dots k$ *respectively* $(k \geq 1)$. *Two elements of distinct subsets are incomparable. Both orders on* $T_C$ *and* $T_R$ *are denoted by* $\leq$ *(*$x \leq y$ *means that* $x$ *is a subtype of* $y$*).* $\mathcal{I}$ *is the set of* individual markers. $T_C$, $T_R$ *and* $\mathcal{I}$ *are pairwise disjoint.* $\tau$ *is a mapping from* $\mathcal{I}$ *to* $T_C$. *We denote by* $*$ *the* generic marker*, where* $* \notin \mathcal{I}$. *A partial order on* $\mathcal{I} \cup \{*\}$ *considers elements of* $\mathcal{I}$ *as pairwise incomparable, and* $*$ *as its greatest element.*

All partial orders will be denoted by $\leq$, and, if needed, indexed by the name of the set on which they are defined. In this paper the intuitive meaning of $x \leq y$ is always *x is a specialization of y*. Fig. 1 partially describes a support, which will be used in further examples.

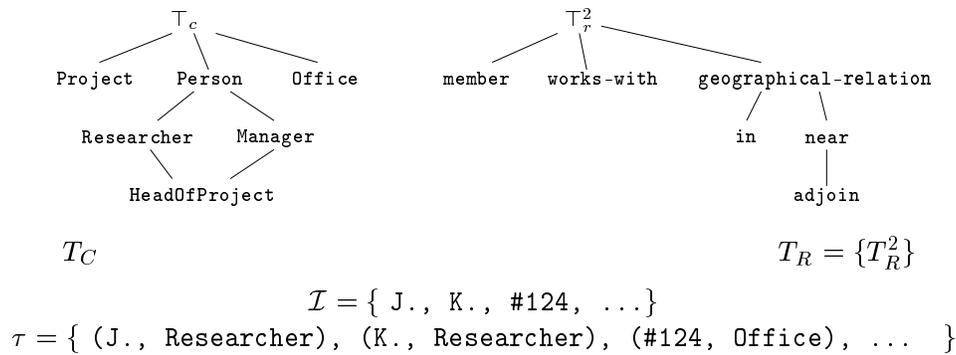

$$\mathcal{I} = \{ \text{ J., K., \#124, } \dots \}$$
$$\tau = \{ \text{ (J., Researcher), (K., Researcher), (\#124, Office), } \dots \}$$

Figure 1: Support

**Definition 2 (Simple Graph)** *A simple graph* $G$*, defined over a support* $\mathcal{S}$*, is a finite bipartite multigraph* $(V = (V_C, V_R), U, \lambda)$. $V_C$ *and* $V_R$ *are the node sets, respectively of* concept nodes *and of* relation nodes. $U$ *is the multiset of* edges. *Edges incident on a relation node are totally ordered; they are numbered from 1 to the degree of the node. An edge numbered* $i$ *between a relation node* $r$ *and a concept node* $c$ *is denoted by* $(r, i, c)$ *and identifies a unique element of* $U$. $U$ *will also be seen as a set of such triples.*

*Each node has a label given by the mapping* $\lambda$. *A relation node* $r$ *is labelled by* type$(r)$*, an element of* $T_R$*, called its* type*, and the degree of* $r$ *must be equal to the arity of* type$(r)$. *Thus, if* $r \in T_R^k$, $|\{(r, i, c) | (r, i, c) \in U\}| = k$ *and* $\{i | (r, i, c) \in U\} = \{1, \dots k\}$. *A concept* node $c$ is labelled by a pair $(\text{type}(c), \text{marker}(c))$, where $\text{type}(c)$ is an element of $T_C$, called its type, and $\text{marker}(c)$ is an element of $\mathcal{I} \cup \{*\}$, called its marker. If $marker(c)$ is an individual marker $m$, then $type(c) = \tau(m)$.

A concept node with a generic marker is called a *generic node* (it refers to an unspecified entity of a certain type); otherwise it is called an *individual node* (it refers to a specific individual defined in the support). We will adopt the following classical conventions about SGs. In the drawing of a SG, concept nodes are represented by rectangles and relation nodes by ovals. In textual notation, rectangles are replaced by [] and ovals by (). Generic markers are omitted. Thus a generic concept label $(t, *)$ is simply noted $t$. An individual concept label $(t, m)$ is noted $t : m$. When in our examples we use binary relations, we may replace numbers on edges by directed edges: a binary relation node is then incident to exactly one incoming and one outgoing edge. Fig. 2 shows two (connected) simple graphs $G$ and $Q$ assumed to be defined over the support of Fig. 1.

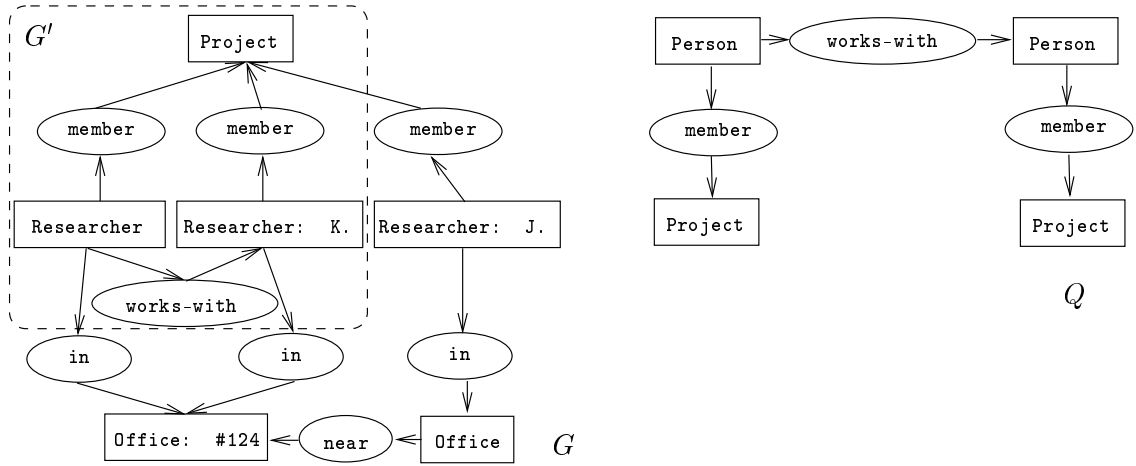

Figure 2: Simple graphs.

The elementary reasoning operation, projection, is a kind of graph homomorphism that preserves the partial order defined on labels. Let us first precise this order for concept node labels. We have defined the following partial order on the marker set $\mathcal{I} \cup \{*\}$: $*$ is the greatest element (for all $m \in \mathcal{I}$, $m \leq *$) and elements of $\mathcal{I}$ are pairwise non comparable. Then the *partial order on concept node labels* is the product of the partial orders on $T_C$ and $I \cup \{*\}$, i.e. $(t, m) \leq (t', m')$ iff $t \leq t'$ and $m \leq m'$. In other words, a concept label $(t, m)$ is more specific than a concept label $(t', m')$ if $t$ is a subtype of $t'$ and, if $m' = *$, then $m$ can be any marker, otherwise $m$ must be equal to $m'$.

**Definition 3 (Projection)** *Let $Q$ and $G$ be two SGs defined on a support $\mathcal{S}$. A projection from $Q$ into $G$ is a mapping $\pi$ from $V_C(Q)$ to $V_C(G)$ and from $V_R(Q)$ to $V_R(G)$ which preserves edges (it is a bipartite graph homomorphism) as well as their numbering, and may specialize concept and relation node labels:*

*1. $\forall (r, i, c) \in U(Q), (\pi(r), i, \pi(c)) \in U(G)$;*

2. $\forall x \in V(Q), \ \lambda(\pi(x)) \leq \lambda(x)$

We note $Q \geq G$ ($Q$ subsumes $G$) if there exists a projection from $Q$ into $G$. Typically, $Q$ represents a query, $G$ a fact, and projections from $Q$ to $G$ define answers to $Q$. In Fig. 2, suppose $Researcher \leq Person$, then there is one projection from $Q$ into $G$. The image of $Q$ by this projection is the subgraph $G'$ of $G$.

## 2.2 Relationships with FOL

The semantics $\Phi$ maps SGs to the existential conjunctive and positive fragment of FOL. Given a support $\mathcal{S}$, a constant is assigned to each individual marker and an $n$-adic (resp. a unary) predicate is assigned to each $n$-adic relation (resp. concept) type. For simplicity, we consider that each constant or predicate has the same name as the associated element of the support. A set of formulas $\Phi(\mathcal{S})$ is assigned to any support $\mathcal{S}$, translating partial orders on types. More specifically, for all distinct types $t_1$ and $t_2$ such that $t_2 \leq t_1$, one has the formula $\forall x_1...x_p(t_2(x_1, \ ..., x_p) \to t_1(x_1, \ ..., x_p))$, where $p = 1$ for concept types, and $p$ is otherwise the arity of the relation type. Given any SG $G$, a formula $\Phi(G)$ is built as follows. A term is assigned to each concept node: a distinct variable for each generic node, and the constant corresponding to its marker otherwise. Then an atom $t(c)$ (resp. $t(c_1, \ ..., c_k)$) is associated to each concept node (resp. relation node $r$ of arity $k$), where $t$ is the type of the node, and $c$ (resp. $c_i$) is the term assigned to this node (resp. assigned to the $i$th neighbour of $r$). Let $\alpha(G)$ be the conjunction of these atoms. $\Phi(G)$ is the existential closure of $\alpha(G)$. E.g. the formula assigned to the subgraph $G'$ in Fig. 2 is $\exists x \exists y (Researcher(x) \wedge Project(y) \wedge Researcher(K) \wedge member(x, y) \wedge member(K, y) \wedge works\text{-}with(x, K))$.

Projection is sound and complete w.r.t. the semantics $\Phi$, up to a normality condition for completeness; the *normal form* of a SG $G$ is the SG $nf(G)$ obtained by merging concept nodes having the same individual marker. This SG always exists (and is computable in linear time with a naive algorithm). Figure 3 shows a counter-example to projection completeness when SGs are not in normal form: $G$ for instance does not project to $H$, even if both SGs have the same logical semantics, but it projects to $nf(H)$. A SG in normal form is said to be *normal*.

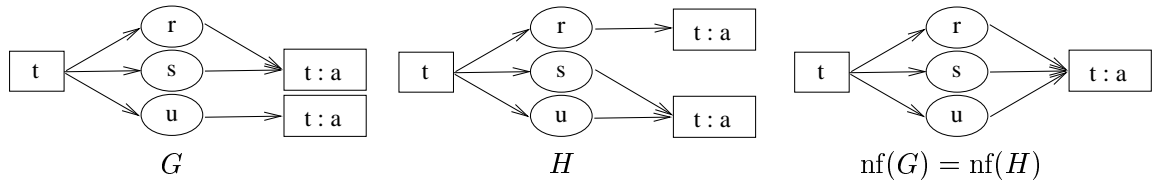

$$\Phi(G) = \Phi(H) = \exists x t(x) \wedge t(a) \wedge t(a) \wedge r(x, \ a) \wedge s(x, \ a) \wedge u(x, \ a)$$

$G$ and $H$ have same logical translation (thus same normal form) but they are incomparable by projection.

Figure 3: The need for normal forms

**Theorem 1 (Chein & Mugnier, 1992; Gosh & Wuwongse, 1995)** *Let $Q$ and $G$ be two SGs defined on a support $\mathcal{S}$. Then $Q \geq nf(G)$ if and only if $\Phi(\mathcal{S}), \Phi(G) \models \Phi(Q)$.*

We claimed in the introduction that SGs are equivalent to the positive, conjunctive and existential fragment of FOL without functions (let us denote it by $FOL(\wedge, \exists)$). One embedding is immediate (from FOL to SGs), but requires the definition of a support that does not add anything to the semantics of the involved graphs. A *flat support* is a support whose translation by $\Phi$ is empty, *i.e.* where all distinct types are non comparable. If $V$ is the vocabulary (constants and predicates) for a set of formulas, we consider the flat support $\mathcal{S}_f(V) = (T_C, T_R, \mathcal{I})$ where $T_C$ is restricted to the element $\top_C$, the relation types of $T_R^i$ are the predicates of arity $i$ in $V$, and the individual markers of $\mathcal{I}$ are the constants in $V$.

**Property 1 (Embedding $FOL(\wedge, \exists)$ into $\mathcal{SG}$)** *There is a bijection $f2g$ mapping the set of $FOL(\wedge, \exists)$ formulas over a vocabulary $V$ to the set of normal SGs defined on the flat support $\mathcal{S}_f(V)$ such that, for any two formulas $g$ and $h$, $g \models h$ iff there is a projection from $f2g(h)$ into $f2g(g)$.*

*Proof:* Let $f$ be a $FOL(\wedge, \exists)$ formula over a vocabulary $V$. The SG $f2g(f)$ defined on the support $\mathcal{S}_f(V)$ is built as follows: to each term of $f$ we associate a concept node typed $\top_C$ (generic if the term is a variable, individual with a marker $c$ if the term is the constant $c$), and to each atom $t(x_1, \ldots, x_q)$ we associate a relation node $r$ typed $t$, such that, for $1 \le i \le q$, the $i$th neighbor of $r$ is the concept node associated to $x_i$.

The mapping $f2g$ is clearly injective, *i.e.* it maps different formulas (not identical up to variable renaming) to different SGs (not identical up to an isomorphism). Moreover, it is a bijection if we restrict SGs to those in normal form.

Let us now consider the $FOL(\wedge, \exists)$ formula $f = \exists \vec{x_i}(\alpha(\vec{x_i}))$ (where $\alpha(\vec{x_i})$ is a conjunction of atoms whose variables belong to $\vec{x_i}$). The $\top_C$-enriched formula of $f$ is the formula $te(f) = \exists \vec{x_i}(\alpha(\vec{x_i}) \wedge \beta(\vec{x_i}))$ where $\beta(\vec{x_i})$ is the conjunction of the atoms $\top_C(x)$, for every term $x$ in $f$. We now prove the property by pointing out that, 1) for $f$ and $g$ two $FOL(\wedge, \exists)$ formulas, $f \models g$ iff $te(f) \models te(g)$, and 2) for any $FOL(\wedge, \exists)$ formula $f$, $te(f) = \Phi(f2g(f))$, and conclude using Th. 1. □

For the other direction, the apparent problem is that formulas assigned to the support by $\Phi$ are universally quantified and are used in the deduction process. However, we can do without them, by encoding the partial order on types directly in the SGs.

**Property 2 (Embedding $\mathcal{SG}$ into $FOL(\wedge, \exists)$)** *There is an injective application $g2f$ mapping the set of normal SGs defined on a support $\mathcal{S}$ to the set of $FOL(\wedge, \exists)$ formulas such that, for any two SGs $G$ and $H$ defined on $\mathcal{S}$, there is a projection from $H$ into $G$ iff $g2f(G) \models g2f(H)$.*

*Proof:* Let $G$ be a graph defined on a support $\mathcal{S}$. The expansion of $G$, $exp(G)$, is the SG defined on the flat support $\mathcal{S}_f(V)$ (where $V$ is the vocabulary for the formulas of $\Phi(\mathcal{S})$), built as follows: 1) for every concept node $x = [t : m]$ of $G$, $exp(G)$ contains an associated concept node $x' = [\top_C : m]$, s.t., for each concept type $t' \in \mathcal{S}$ greater or equal to $t$, a unary relation node of type $t'$ linked to $x'$ and 2) for every relation node $x$ of $G$ (of type $r$ and arity $k$), for every relation type $r' \in \mathcal{S}$ s.t. $r \le r'$, we add in $exp(G)$ a relation node typed $r'$ with same neighbors as $x$. We now define the application $g2f$ as $\Phi \circ exp$, and conclude using Th. 1, noticing that $\Phi(\mathcal{S}_f(V)) = \emptyset$.

□

Using similar transformations, a close relationship to the problem of *query containment* studied in the database field has been shown: checking query containment for non recursive conjunctive queries is equivalent to checking projection between SGs (Chein et al., 1998; Mugnier, 2000).

## 2.3 The Deduction Problem: Computational Complexity

For the sake of brevity, we consider in what follows that SGs are given in normal form, and put into normal form if needed after a modification. And, since a SG does not need to be a connected graph, we conflate a set of SGs with the SG obtained by performing the disjoint union of its elements. In the following definition, for instance, the SG $\mathcal{G}$ represents a set of SGs.

**Definition 4 ($\mathcal{SG}$-DEDUCTION)** *Let $\mathcal{G}$ and $Q$ be two SGs defined on a support $\mathcal{S}$. $Q$ can be deduced from $\mathcal{G}$ if $Q \geq \mathcal{G}$.*

Chein and Mugnier (1992) have shown that projection checking is NP-complete with a reduction from CLIQUE. Equivalence with CSP (satisfiability of a constraint network) was also used later (Feder & Vardi, 1993; Mugnier & Chein, 1996), and independently in a very similar model by Rudolf (1998) (see part 8 of this paper). We give below another proof of this result based on a reduction from 3-SAT. Though more complicated than the previous ones, this reduction is the basis for other reductions presented later in this paper.

The following theorem keeps into account the complexity of concept and relation type checking, though in this paper this test can obviously be performed in polynomial time since concept and relation types are only labels partially ordered in a hierarchy.

**Theorem 2 (Chein & Mugnier, 1992)** *$\mathcal{SG}$-DEDUCTION is a NP-complete problem, iff type checking in $\mathcal{S}$ belongs to NP.*

*Proof:* First see that if type checking is in NP, then $\mathcal{SG}$-DEDUCTION is also in NP: a projection, enriched by certificates for all type checks used, is a polynomial certificate. The reciprocal is obviously true. We now show that, even if type checking can be done in $\Theta(1)$, $\mathcal{SG}$-DEDUCTION is NP-complete.

Let us now build a reduction from 3-SAT. The input of 3-SAT is a formula $\mathcal{F}$ in 3-conjunctive normal form (3-CNF), i.e. a conjunction of disjunctions (clauses), each with at most three literals, and the question is whether there is a truth assignment of the variables of $\mathcal{F}$ such that $\mathcal{F}$ is true. Notice the classical 3-SAT problem considers clauses with exactly three literals, but for further proofs we prefer to use the above variant.

Let $\mathcal{F} = C_1 \wedge \ldots \wedge C_k$ be an instance of 3-SAT. W.l.o.g. we suppose that each variable appears at most once in a clause. Let us create four concept types for each variable $x$: $x$, $xf$, $xt$ and $xv$. We also create one relation type $C_i$ for each clause $C_i$, and a relation type $val$. Each concept type $xv$ is greater than $xt$ and $xf$, these are the only possible comparisons between distinct types.

We build the graph $G(\mathcal{F})$ as follows: for every variable $x$ in $\mathcal{F}$, we have three concept nodes $[x]$, $[xt]$ and $[xf]$ in $G(\mathcal{F})$ and two relation nodes typed $val$ linking the first to the latter ones (intuitively, it means that the variable $x$ can be valuated by $true$ or $false$). Let us say that the truth value `true` (resp. `false`) is associated with `[xt]` (resp. `[xf]`). Then for every clause $C_i = (l_x \vee l_y \vee l_z)$ in $\mathcal{F}$ (where $l_x$, $l_y$ and $l_z$ are literals over variables $x$, $y$ and $z$), we add the 7 relation nodes typed $C_i$, having as first argument `[xt]` or `[xf]`, as second argument `[yt]` or `[yf]`, and as third argument `[zt]` or `[zf]`, that correspond to an evaluation of the clause to true (more precisely, if we replace in the clause $C_i$ each positive (resp. negative) literal $l_j$, $1 \leq j \leq 3$, by the truth value (resp. the negation of the truth value) associated with the $jth$ neighbor of the relation node, $C_i$ is evaluated to true).

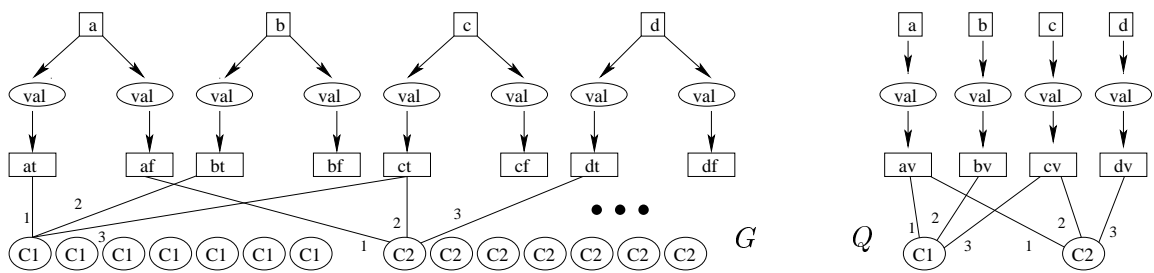

Figure 4: Example of transformation from 3-SAT to PROJECTION

For clauses restricted to $(l_x \vee l_y)$ or $(l_x)$, we proceed similarly, adding 3 binary relation nodes or one unary relation node. Note that having $k$-clauses, where $k$ is a constant, is of primary importance to have a polynomial transformation, since we obtain $2^k - 1$ relation nodes for each clause.

In the graph $Q(\mathcal{F})$, two concept nodes `[x]` and `[xv]` are created for each variable $x$ and a binary relation node (`val`) links `[x]` to `[xv]`. For each clause $C_i = (l_x \vee l_y \vee l_z)$, there is one relation (`C_i`) linked to `[xv]`, `[yv]` and `[zv]` (and similarly for clauses with one or two literals). $Q(\mathcal{F})$ represents the question "is there a valuation of the variables such that all clauses evaluate to `true`?"

This transformation from the 3-SAT formula $(a \vee b \vee \neg c) \wedge (\neg a \vee c \vee \neg d)$ is illustrated in Fig. 4. In the graph $G$, not all edges issued from the clauses have been drawn, for the sake of readability. It is immediate to check that, for a formula $\mathcal{F}$, there is a valuation of its variables such that each clause is evaluated to true if and only if $Q(\mathcal{F})$ can be projected into $G(\mathcal{F})$. □

## 2.4 A Note on Redundancy

Note that the subsumption relation induced by projection over SGs is a quasi-order, but not an order: it is a reflexive and transitive but not anti-symmetric relation. Two SGs are said to be *equivalent* if they project to each other. A SG is said to be *redundant* if it is equivalent to one of its strict subgraphs (i.e. a subgraph not equal to $G$ itself), otherwise it is said to be *irredundant*.

**Theorem 3 (Chein & Mugnier, 1992)** *Redundancy checking is an NP-complete problem. Each equivalence class admits a unique (up to isomorphism) irredundant graph.*

The *irredundant form* of a SG $G$ is an irredundant subgraph of $G$ equivalent to it (when $G$ is irredundant, this graph is $G$ itself, otherwise there may be several such subgraphs, but they are all isomorphic).

## 3. The $\mathcal{SG}$ Family: Extensions of Simple Conceptual Graphs

This section is devoted to an overview of the different models composing the $\mathcal{SG}$ family. We will first outline the main motivations for our graph-based approach of knowledge representation.

### 3.1 Knowledge Representation and Reasonings with Graphs

**A modeling viewpoint**  From a modeling viewpoint, we see two essential properties in the simple graph model. The *objects*, simple graphs, are easily understandable by an end-user (a knowledge engineer or even an expert). And *reasonings* are easily understandable too, for two reasons: projection is a graph matching operation, thus easily interpretable and visualisable; and the same language is used at interface and operational levels.

Although there is a gap between the theoretical foundations studied here and a language usable in real applications, we would like to briefly mention two projects in which these properties have been exhibited. The first one is an experiment in document retrieval done by Genest (2000). In this work, conceptual graphs are used to define a language for indexing and querying documents. Concept types are taken from the thesaurus of RAMEAU (about 400 000 types), a documentary language used in most french public and universitary libraries. The experiment proved the feasibility of the proposed system (w.r.t. computing time) and an improved relevance w.r.t. to the existing system based upon RAMEAU, mainly due to the use of semantic relations instead of keywords only. One side effect was also to prove the interest of simple graphs from a modeling viewpoint. Indeed, their graphical properties enabled to build an indexing/querying tool that was considered as easy to use for the indexers. The users were master humanities students, not aware from conceptual graphs neither from RAMEAU; with the software and an indexing guide, they became quickly able to build indexations, that were considered of high quality by a senior librarian.

The second project takes place in knowledge engineering (Bos, Botella, & Vanheeghe, 1997). Its purpose is the construction of tools for modeling and simulating human organizations, as emergency procedures for instance. One main difficulty in knowledge engineering is to validate a modeling, i.e. to check that the expert reasoning is correctly modeled. This validation is usually done when the design is achieved, here by simulating the constructed modeling of the organization. At this final stage, modifications are very costly. The key idea of the project is to overcome this difficulty by giving the expert the ability to use simulation inside the design cycle as a mean of enriching and building his modeling. This implies that the chosen modeling language enables the expert to follow reasonings step by step, directly on his own modelization. It was decided to build such a language upon conceptual graphs. General conceptual graphs equivalent to FOL were not considered as good candidates because they are indeed a diagrammatic system of logic that is not at the expert level. Instead, the language was grounded upon simple graphs and extensions (such as nestings of graphs) keeping their readability. Operations mixed simple graph deduction (i.e. projection) with non declarative procedures. First experiments were conclusive.

**A computational viewpoint**  From a computational viewpoint, we think that graph-based reasonings, benefitting from graph-theoretical results, can bring an interesting perspective to logic programming. By example, the equivalence between SG projection and deduction in FOL($\land, \exists$) can be seen as an alternative version of the *homomorphism theorem* (Chandra & Merlin, 1977), considered as fundamental for database queries optimization (Abiteboul, Hull, & Vianu, 1995). Other results are obtained from constraint programming. The strong equivalence between $\mathcal{SG}$-DEDUCTION and the CONSTRAINT SATISFACTION PROBLEM (see Sect. 8, where the transformations used keep all solutions and preserve the structure of the constraint network in the query) allows to translate the results obtained in this latter community (by example, tractable cases based upon the structure of the graph, Gottlob, Leone, & Scarcello, 1999), first to $\mathcal{SG}$-DEDUCTION, then to deduction in FOL($\land, \exists$).

The graph structure can also be used to develop efficient algorithms in more general models of the $\mathcal{SG}$ family: in the model we call $\mathcal{SR}$ (see below), Coulondre and Salvat (1998) use the graph-based notion of *piece* to build an efficient backward-chaining algorithm. To enhance the forward-chaining algorithm used in the more general models of the $\mathcal{SG}$ family, Baget (2001) expresses dependencies between rules and constraints in terms of a graph homomorphism.

Our aim is thus to build formal extensions of simple conceptual graphs, keeping readability of objects as well as reasonings, and preferably, logically founded. The $\mathcal{SG}$ family is a first step in this direction.

## 3.2 An Overview of the $\mathcal{SG}$ Family

Let us now informally present the $\mathcal{SG}$ family. The generic problem to be solved, DEDUCTION, asks, given a knowledge base (KB) $\mathcal{K}$ and a simple graph $Q$, whether $Q$ can be deduced from $\mathcal{K}$. According to the kinds of objects composing $\mathcal{K}$, one obtains the different members of the family. In the basic model $\mathcal{SG}$, $\mathcal{K}$ is composed of a set $\mathcal{G}$ of simple graphs representing facts, and solving DEDUCTION amounts to check whether there is a projection from $Q$ into $\mathcal{G}$. Rules and constraints are more complex objects based upon simple graphs, and operations dealing with these objects are based upon projection.

Throughout this section, we will use examples inspired from a modelization of a knowledge acquisition case study, called SYSIPHUS-I: it describes a resource allocation problem, where the aim is to assign offices to persons of a research group while fulfilling some constraints (Baget et al., 1999).

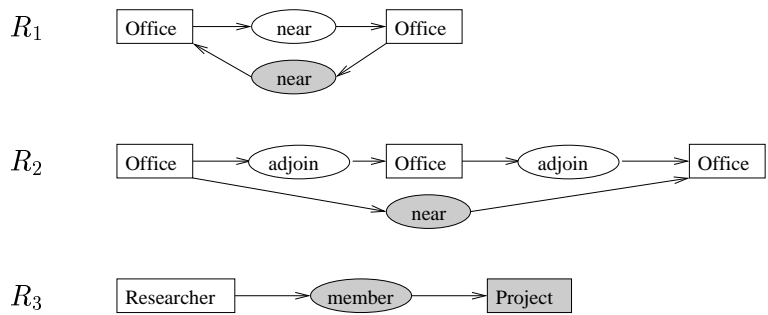

Figure 5: Rules

**Rules** A *rule* expresses knowledge of form "if $A$ is present then $B$ can be added". It is encoded into a simple graph provided with two colors, the first color subgraph defining the hypothesis and the second color the conclusion. In drawings, we represent the hypothesis by white nodes, and the conclusion by gray ones. Figure 5 shows three rules. $R_1$ and $R_2$ represent knowledge about the *near* relation, supposed to be defined between offices only. $R_1$ expresses that the relation *near* is symmetrical ("if an office $x$ is *near* an office $y$, then $y$ is *near* $x$"), $R_2$ that " if an office $x$ *adjoins* an office $y$ that *adjoins* an office $z$ then $x$ is *near* $z$". The rule $R_3$ says that every researcher is member of a project ("if there is a researcher $x$, there is a project of which $x$ is a member").

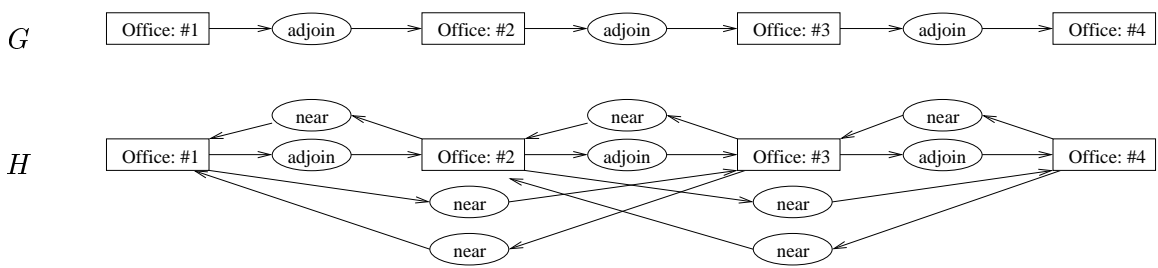

Figure 6: Rule applications

Rules are used to enrich facts: if the hypothesis of a rule can be projected into a SG, then the rule is applicable to this SG, and its conclusion can be added to the SG according to the projection. Notice that each projection of a same rule to a SG defines a different way of applying this rule and is likely to add new information to the SG. Consider for instance the SG $G$ of Fig. 6, which describes spatial information about offices, and rules of Figure 5. $R_1$ is applicable (since *adjoin* $\leq$ *near*), and so is $R_2$. Let us consider $R_2$. There are two ways of applying this rule, depending on whether its hypothesis is mapped onto the path `[Office:#1]->(adjoin)->[Office:#2]->(adjoin)->[Office:#3]` or onto the path `[Office:#2]->(adjoin)->[Office: #3]->(adjoin)->[Office:#4]`. In the first case for instance, a relation node `(near)` with predecessor `[Office:#1]` and successor `[Office:#3]` is added to the SG. Notice that in this example, applying all rules in all possible ways as long as they add new information is a finite process (leading to the graph $H$ of Figure 6) but it is not true in general.

When the KB is composed of a set of facts $\mathcal{G}$ and a set of rules $\mathcal{R}$, the DEDUCTION problem asks whether there is a sequence of rule applications enriching the facts such that the goal $Q$ can be reached, i.e. leading to a graph into which the SG $Q$ can be projected. E.g. consider the fact $G$ of Figure 6, and let $Q$ be the SG `[Office:#4]->(near)->[Office]` ("is #4 near an office?"). $Q$ does not project into $\mathcal{G}$, but applying the rules, one adds the information `[Office:#4]->(near)->[Office:#3]` (also `[Office:#4]->(near)->[Office:#2]`), thus answering $Q$.

**Constraints**  A *constraint* can be positive or negative, expressing knowledge of form "if $A$ holds, so must $B$", or "if $A$ holds, $B$ must not". It is also a bicolored simple graph: the first color defines the condition part (or *trigger*), and the second color the mandatory (or forbidden) part. A SG $G$ satisfies a positive constraint $C$ if *each* projection from the condition part of $C$ into $G$ can be extended to a projection of the whole $C$. And $G$ satisfies a negative constraint if *no* projection of the condition of $C$ into $G$ can be extended to a projection of the whole $C$. Fig. 7 shows two constraints. The negative constraint $C_1$

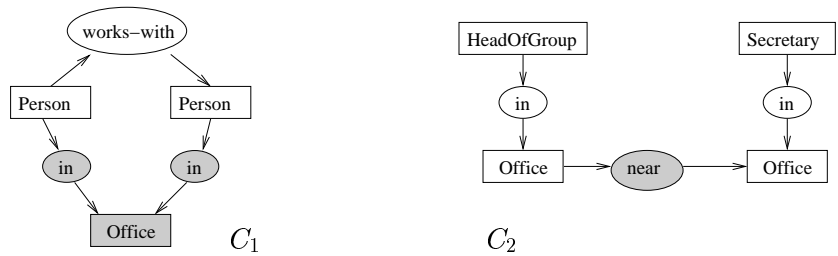

PSfrag replacements

$C_1$             $C_2$

Figure 7: Constraints

expresses that "two persons working together should not share an office". The SG $G$ of Fig 2 does not satisfy this constraint because "there is a researcher who works with researcher K" (projection of the condition part of $C1$) "and they share office #124" (extension of the projection to a projection of the whole $C1$). The positive constraint $C_2$ expresses that the office of a head of group must be near the offices of all secretaries.

When the KB is composed of a set of facts $\mathcal{G}$ and a set of constraints $\mathcal{C}$, the role of constraints is to define the consistency of the base, i.e. of $\mathcal{G}$. The base is said to be consistent if all constraints are satisfied. Provided that the base is consistent, deduction is done as in $\mathcal{SG}$. Even if they are both bicolored graphs, constraints are not to be confused with rules. Consider for instance the bicolored graph $R_3$ of Figure 5: as a rule, it says that every researcher is a member of a project. Take the fact $\mathcal{G} = $ [Researcher:K.] and the query $Q=$[Researcher:K.]->(member)->[Project] ("is K. member of a project?"). If $Q$ is asked on $\mathcal{K} = (\mathcal{G}, \mathcal{R} = \{R3\})$, the answer is "yes". Now, see $R_3$ as a positive constraint $C$. It says that every researcher must be a member of a project. $\mathcal{K} = (\mathcal{G}, \mathcal{C} = \{C\})$ is inconsistent, thus nothing can be deduced from it, including $Q$. The KB has to be repaired first.

**Combining rules and constraints** Let us combine rules and constraints in reasoning. We distinguish now between two kinds of rules: *inference rules* and *evolution rules*.

Inference rules represent implicit knowledge that is made explicit by rule applications. This is the case for rules seen above (Figure 5). Facts and inference rules can be seen as describing a world, and applying a rule modifies the explicit description of the world (the facts). Now, if we consider a KB composed of a set of facts $\mathcal{G}$, a set of inference rules $\mathcal{R}$, and a set of constraints $\mathcal{C}$, the notion of consistency has to take rules into account. For instance, add to the SGs $G$ and $H$ of Figure 6 the following information about office assignments: [HeadOfgroup:L.]->(in)->[Office:#1], [Secretary:H.]->(in)->[Office:#2] and [Secretary:P.]->(in)->[Office:#3]. Let $G'$ and $H'$ be the (normal) SGs obtained. Consider the world composed of the SG $G'$, inference rules $\{R_1, R_2\}$ of Figure 5, and the positive constraint $C_2$ of Figure 7. The SG $G'$ alone does not satisfy the constraint $C2$ (because "the head of group L. is in office #1, and the secretary P. is in office #3", but it does not hold that "#1 is near #3"). But after a certain number of rule applications, it does. Thus the KB is said to be consistent. In this case it is easy to define and check consistency because the world description can be completely explicited by a finite SG (the graph $H'$, said to be *full* w.r.t. $\mathcal{R}$), thus it suffices to check that this graph is consistent. In general case, consistency relies on whether each "constraint violation" can be repaired by rule applications, as will be formally defined later. As with simpler worlds described by facts only, deduction is not possible on inconsistent knowledge bases.

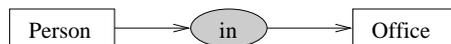

Figure 8: A rule

Evolution rules represent possible actions leading from one world to another one. E.g. consider the colored graph of Figure 8. As an inference rule, it would allow to deduce that all persons are in all offices. As an evolution rule, it says that "when there are a person and an office, a possible action is to assign this office to that person". Consider a KB composed of a set of facts $\mathcal{G}$, a set of evolution rules $\mathcal{E}$, and a set of constraints $\mathcal{C}$. Facts describe an initial world; evolution rules represent possible transitions from one world to other worlds; constraints define consistency of each world; a successor of a consistent world is obtained by an evolution rule application; given a SG $Q$, the deduction problem asks whether there is a path of consistent worlds evolving from the initial one to a world satisfying $Q$.

The most general model of the $\mathcal{SG}$ family considers both kinds of rules, $i.e.$ a set $\mathcal{R}$ of inference rules, and a set $\mathcal{E}$ of evolution rules. In the particular case of the SYSIPHUS-I modelization, $\mathcal{G}$ and $\mathcal{R}$ describe the initial information about office locations, persons and the group organization. $\mathcal{R}$ also encodes general knowledge (such as properties of the $dif$ relation put between two concept nodes representing distinct entities). $\mathcal{C}$ represents obligations and interdictions defining what acceptable assignments are (including cardinality constraints such as "a person cannot be $in$ several offices" or "a large office cannot contain more than two persons", using the $dif$ relation). $\mathcal{E}$ consists of one evolution rule whose result is to place a person into an office (it could also be composed of several rules considering specific preconditions before trying an assignment). The goal represents a situation where each person of the group has an office. A solution to the problem is a world obtained from the initial one by a sequence of office assignments, where each person has an office, while satisfying the allocation constraints.

### 3.3 The $\mathcal{SG}$ Family

Let us now specify definitions and notations concerning the $\mathcal{SG}$ family.

**Definition 5 (colored SGs)** A colored simple graph *is a pair* $K = (G, \rho)$ *where $G$ is a SG and $\rho$ is a mapping from* $V(G)$ *into* $\{0, 1\}$. *The number associated to a node is called the* color *of the node. We denote by* $K_{(i)}$ *the subgraph of $G$ induced by $i$-colored nodes. The subgraph* $K_{(0)}$ *must form a SG (i.e. the neighbors of a relation node of $K_{(0)}$ must also belong to* $K_{(0)}$).

The latter condition ($K_{(0)}$ must form a SG) is necessary as soon as we consider rules as colored SGs: should a rule not satisfy this condition, its application on a SG could generate a graph that is not a SG.

A KB is denoted by $\mathcal{K} = (\mathcal{G}, \mathcal{R}, \mathcal{E}, \mathcal{C})$, where $\mathcal{G}$ is a set of simple graphs representing facts, $\mathcal{R}$, $\mathcal{E}$ and $\mathcal{C}$ are three sets of colored simple graphs respectively representing *inference rules*, *evolution rules*, and *constraints* (positive ones in $\mathcal{C}^+$, negative ones in $\mathcal{C}^-$). Given a KB $\mathcal{K}$ and a goal $Q$, the *deduction problem* asks whether $Q$ can be deduced from $\mathcal{K}$ (we note $Q \geq \mathcal{K}$). If we impose some of the sets $\mathcal{R}$, $\mathcal{E}$ or $\mathcal{C}$ to be empty, one obtains specific reasoning models. Note that in the absence of constraints ($\mathcal{C} = \emptyset$), inference and evolution rules have the same behavior, thus $\mathcal{R}$ and $\mathcal{E}$ can be confused. The $\mathcal{SG}$ family is then composed of the six following models.

- the $\mathcal{SG}$ model for $\mathcal{K} = (\mathcal{G}, \ \emptyset, \ \emptyset, \ \emptyset)$

- the $\mathcal{SR}$ model for $\mathcal{K} = (\mathcal{G}, \ \mathcal{R}, \ \mathcal{E}, \ \emptyset)$

- the $\mathcal{SGC}$ model for $\mathcal{K} = (\mathcal{G}, \ \emptyset, \ \emptyset, \ \mathcal{C})$

- the $\mathcal{SRC}$ model for $\mathcal{K} = (\mathcal{G}, \ \mathcal{R}, \ \emptyset, \ \mathcal{C})$

- the $\mathcal{SEC}$ model for $\mathcal{K} = (\mathcal{G}, \ \emptyset, \ \mathcal{E}, \ \mathcal{C})$

- the $\mathcal{SREC}$ model for $\mathcal{K} = (\mathcal{G}, \ \mathcal{R}, \ \mathcal{E}, \ \mathcal{C})$

Since a fact has the same semantics as a rule with an empty hypothesis, the set $\mathcal{G}$ is used in models names only when both rule sets $\mathcal{R}$ and $\mathcal{E}$ are empty. The hierarchy of these models is represented in Fig. 9. It highlights the decidability properties and the complexity of the associated deduction problem. Notice we divide non decidable problems into *semi-decidable* and *truly undecidable* problems. In the first case, an answer can be computed in finite time for all positive instances but not for all negative ones. In the second case, there is no finite procedure, neither for all positive instances, nor for all negative ones.

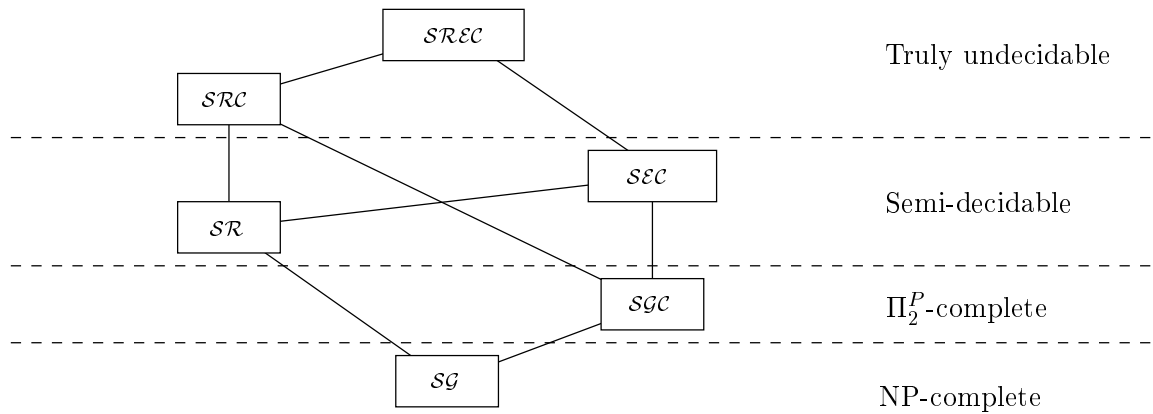

Figure 9: The $\mathcal{SG}$ family: models and complexity of the associated deduction problem

## 4. SGs and Rules: the $\mathcal{SR}$ Model

A *simple graph rule* (SG rule) embeds knowledge of form "if $A$ then $B$". The following definition as a colored SG is equivalent to the more traditional definition of a rule as an object composed of two SGs related with coreference links used by Gosh and Wuwongse (1995), or Salvat and Mugnier (1996).

### 4.1 Definitions and Notations

**Definition 6 (SG Rules)** *A simple graph rule $R$ is a colored SG. $R_{(0)}$ is called its* hypothesis, *and $R_{(1)}$ its* conclusion.

Deduction depends on the notion of a *rule application*: it is a graph transformation based upon projection.

**Definition 7 (Application of a SG Rule)** *Let $G$ be a SG, and $R$ be a rule. $R$ is* applicable *to $G$ if there exists a projection, say $\pi$, from $R_{(0)}$ (the hypothesis of $R$) into $G$. In that case, the result of the* application *of $R$ on $G$ according to $\pi$ is the SG $G'$ obtained by making the disjoint union of $G$ and of a copy of $R_{(1)}$ (the conclusion of $R$), then, for every edge $(r, i, c)$, where $c \in R_{(0)}$ and $r \in R_{(1)}$, adding an edge with the same number between $\pi(c)$ and the copy of $r$. $G'$ is said to be an* immediate $R$-derivation *from $G$.*

A derivation is a (possibly empty) sequence of rule applications:

**Definition 8 (Derivation)** *Let $\mathcal{R}$ be a set of rules, and $G$ be a SG. We call $\mathcal{R}$-derivation from $G$ to $G'$ a sequence of SGs $G = G_0, \ldots, G_k = G'$ such that, for $1 \le i \le k$, $G_i$ is an immediate $R$-derivation from $G_{i-1}$, where $R$ is a rule in $\mathcal{R}$.*

To deduce a SG $Q$, we must be able to derive a SG into which $Q$ can be projected. This notion is captured by the following definition:

**Definition 9 ($\mathcal{SR}$-DEDUCTION)** *Let $\mathcal{K} = (\mathcal{G}, \mathcal{R})$ be a KB and let $Q$ be a SG. $Q$ can be deduced from $\mathcal{K}$ (notation $Q \ge (\mathcal{G}, \mathcal{R})$) if there exists an $\mathcal{R}$-derivation from $\mathcal{G}$ to a SG $H$ such that $Q \ge H$.*

### 4.2 Logical Semantics

The semantics $\Phi$ is extended to translate rules: given a rule $R$, let $R_0$ and $R_1$ be the two SGs respectively corresponding to its hypothesis and its conclusion, i.e. $R_0 = R_{(0)}$ and $R_1$ is the SG obtained from $R_{(1)}$ by adding the neighbors of the relation nodes of $R_{(1)}$ which are concept nodes of $R_{(0)}$. Then $\Phi(R) = \forall x_1 \ldots x_p \ (\alpha(R_0) \to \exists y_1 \ldots y_q \ \alpha(R_1))$ where $\alpha(R_0)$ and $\alpha(R_1)$ are the conjunctions of atoms associated with $R_0$ and $R_1$, $x_1 \ldots x_p$ are the variables of $\alpha(R_0)$ and $y_1 \ldots y_q$ are the variables of $\alpha(R_1)$ that do not appear in $\alpha(R_0)$. For instance, consider the rule $R_3$ in Fig. 5. Then $\Phi(R_3) = \forall x (Researcher(x) \to \exists y(Project(y) \land member(x, y)))$. Should we interpret the colored graph $C_1$ in Fig. 7 as a rule, its formula would be $\Phi(C_1) = \forall x \forall y((Person(x) \land Person(y) \land works\text{-}with(x, y)) \to \exists z(Office(z) \land in(x, z) \land in(y, z)))$. Notice, unlike in clauses, variables proper to the conclusion are existentially quantified.

The following soundness and completeness result is obtained:

**Theorem 4 (Salvat & Mugnier, 1996; Salvat, 1998)** *Let $\mathcal{K} = (\mathcal{G}, \mathcal{R})$ be a KB and $Q$ be a SG. Then $Q \ge (\mathcal{G}, \mathcal{R})$ iff $\Phi(\mathcal{S}), \Phi(\mathcal{G}), \Phi(\mathcal{R}) \vDash \Phi(Q)$.*

Notice this result assumes that graphs are given in normal form, and, if needed, put into their normal form after each rule application.

### 4.3 A Semi-decidable Problem

Coulondre and Salvat (1998) proved that $\mathcal{SR}$-deduction is semi-decidable with a reduction from the THE IMPLICATION PROBLEM FOR TGDs. The reduction given by Baget (2001) (from the HALTING PROBLEM OF A TURING MACHINE) points out that $\mathcal{SR}$-deduction is a computation model. We give here another reduction, from the WORD PROBLEM IN A SEMI-THUE SYSTEM, that we will use as the starting point in the proof of Prop. 10. This reduction is also interesting in itself since it proves that, even when rules are of the form "if *path o* $x_1 \ldots x_k$ *e* then *path o* $y_1 \ldots y_q$ *e*", $\mathcal{SR}$-deduction remains semi-decidable.

**Theorem 5 (Coulondre & Salvat, 1998)** $\mathcal{SR}$-deduction *is semi-decidable.*

*Proof:* First check that $\mathcal{SR}$-deduction is not truly undecidable (*i.e.* there exists an algorithm that can decide in finite time if the answer to the problem is "yes"): when $Q$ can be deduced from $\mathcal{K}$, a breadth-first search of the tree of all derivations from $\mathcal{K}$ provides the answer in finite time.

We then prove that no algorithm is ensured to halt when the answer to the problem is "no". Let us now show that $\mathcal{SR}$-deduction is not decidable by building a reduction from the WORD PROBLEM IN A SEMI-THUE SYSTEM (Thue, 1914). This problem was proven semi-decidable by Post (1947, reduction to his correspondence Problem).

The WORD PROBLEM can be expressed as: let $m$ and $m'$ be two words, and $\Gamma = \{\gamma_1, \ldots, \gamma_k\}$ be a set of rules, each rule $\gamma_i$ being a pair of words $(\alpha_i, \beta_i)$: is there a derivation from $m$ to $m'$? There is an immediate derivation from $m$ to $m'$ (we note $m \rightarrow m'$) if, for some $\gamma_j$, $m = m_1\alpha_j m_2$ and $m' = m_1\beta_j m_2$. A *derivation* from $m$ to $m'$ (we note $m \rightsquigarrow m'$) is a sequence $m = m_0 \rightarrow m_1 \rightarrow \ldots \rightarrow m_p = m'$.

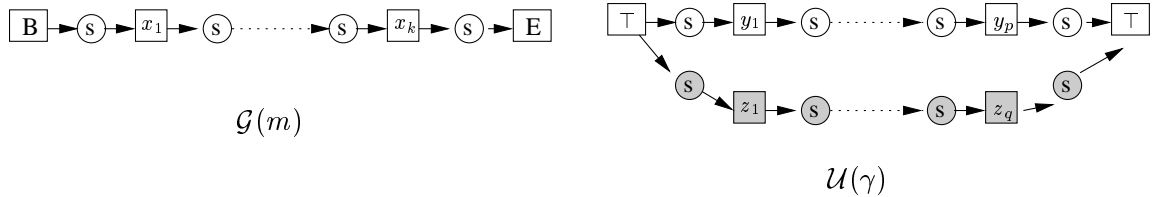

$$\mathcal{G}(m) \qquad\qquad \mathcal{U}(\gamma)$$

Figure 10: Transformation from the WORD PROBLEM into $\mathcal{SR}$-deduction

This problem can easily be expressed in the $\mathcal{SR}$ model. One concept type $\mathtt{x}_i$ is assigned to each letter $x_i$. There are three other concept types: $\mathtt{B}$ (for "begin"), $\mathtt{E}$ (for "end") and $\top$ (for "anything"). $\top$ is the greatest concept type and all other types are pairwise non-comparable. There is one relation type $\mathtt{s}$ (for "has successor"). A word $m = x_1 \ldots x_k$ is associated the graph $\mathcal{G}(m)$, and to any rule $\gamma = (y_1 \ldots y_p, \ z_1 \ldots z_q)$ is associated the graph rule $\mathcal{U}(\gamma)$, as represented in Fig. 10. By a straightforward proof (a recurrence on the smallest derivation length), we obtain that to every path from the node typed $B$ to the node typed $E$ ("begin" to "end") in a graph $\mathcal{R}$-derived from $\mathcal{G}(m)$, corresponds a word (and not a subword) derivable from $m$, and reciprocally. It follows that $m \rightsquigarrow m' \ \Leftrightarrow \ \mathcal{G}(m') \geq (\mathcal{G}(m), \mathcal{U}(\Gamma))$. $\qquad\square$

## 5. SGs and Constraints: the $\mathcal{SGC}$ Model

Let us now introduce *constraints*, which are used to validate knowledge. A knowledge base will be validated if it satisfies every constraint, and no deduction will be allowed unless the KB has been validated: in presence of constraints, deduction is defined only on a *consistent* knowledge base.

### 5.1 Definitions and Some Immediate Properties

**Definition 10 (Constraints)** *A positive (resp. negative) constraint $C$ is a colored SG. $C_{(0)}$ is called the* trigger *of the constraint, $C_{(1)}$ is called its* obligation *(resp. interdiction). A SG $G$ $\pi$-violates a positive (resp. negative) constraint $C$ if $\pi$ is a projection of the trigger of $C$ into the irredundant form of $G$ (resp. into $G$) that cannot be extended (resp. that can be extended) to a projection of $C$ as a whole. $G$ violates $C$ if it $\pi$-violates $C$ for some projection $\pi$. Otherwise, $G$ satisfies $C$.*

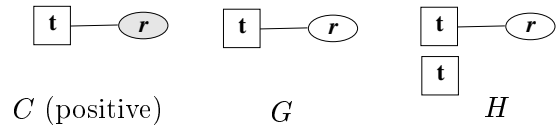

Figure 11: Redundancy and constraint violation

We have to point out the importance of the irredundancy condition on the graph to be validated by positive constraints: should we forget this condition, there may be two equivalent SGs, such that one satisfies a positive constraint and the other does not. Fig. 11 shows an example of such graphs. $G$ satisfies $C$, but the equivalent (redundant) graph $H$, obtained by making the disjoint union of $G$ and the trigger of $C$, does not. To avoid different consistency values for equivalent graphs, we have chosen to define positive constraint satisfaction w.r.t. the irredundant form of a SG. This problem does not occur with negative constraints. Indeed, let $G_1$ and $G_2$ be two equivalent graphs and suppose $G_1$ $\pi$-violates a negative constraint $C$; since there exists a projection from $G_1$ into $G_2$, say $\pi_1$, $\pi_1 \circ \pi$ is a projection from $C$ to $G_2$, thus $G_2$ also violates $C$.

Two constraints $C_1$ and $C_2$ are said to be *equivalent* if any graph that violates $C_1$ also violates $C_2$ and conversely. Any negative constraint is equivalent to the negative constraint obtained by coloring all its nodes by 1. Furthermore, negative constraints are indeed a particular case of positive ones: consider the positive constraint $C'$ obtained from a negative constraint $C$ by coloring all nodes of $C$ by 0, then adding a concept node colored by 1, with type `NotThere`, where `NotThere` is incomparable with all other types and does not appear in any graph of the KB, except in constraints. Then a simple graph $G$ violates the constraint $C$ if and only if it violates $C'$. Positive constraints strictly include negative constraints, in the sense that the associated consistency problems are not in the same complexity class (the proof follows from Th. 8).

**Property 3** *Unless $\Pi_2^P =$ co-NP, positive constraints are a strict generalization of negative ones.*

Since negative constraints are indeed a particular case of positive ones, we will now, unless indicated otherwise, denote by "a set of constraints" a set of positive constraints: some of them can be equivalent to negative ones.

**Definition 11 (Consistency/Deduction in $\mathcal{SGC}$)** *A KB $\mathcal{K} = (\mathcal{G}, \mathcal{C})$ is consistent if $\mathcal{G}$ satisfies all constraints of $\mathcal{C}$. Otherwise, it is said inconsistent. A SG $Q$ can be deduced from $\mathcal{K}$ if $\mathcal{K}$ is consistent and $Q$ can be deduced from $\mathcal{G}$.*

Note that a SG $Q$ that violates a constraint of $\mathcal{K}$ may still be deduced from $\mathcal{K}$. It does not matter since $Q$ is a *partial* representation of knowledge deducible from $\mathcal{K}$.

## 5.2 Relationships with Logics

Deduction in $\mathcal{SGC}$ is essentially non monotonic. Adding information to $\mathcal{G}$ can trigger a new constraint, and thus can create a new violation: since nothing can be deduced from an inconsistent knowledge base, previous deductions are no longer valid. That is why for $\mathcal{SGC}$ and more general models (next sections), it is impossible to obtain results of form "$Q$ can be deduced from the knowledge base $\mathcal{K}$ iff $\Phi(\mathcal{K}) \vDash \Phi(Q)$" as it was the case for $\mathcal{SG}$ and $\mathcal{SR}$.

However, the notion of consistency can be translated into FOL. For negative constraints, the correspondence is immediate, and relies on projection soundness and completeness w.r.t. the semantics $\Phi$ (theorem 1). Intuitively, a SG $G$ violates a negative constraint $C^-$ if and only if the information represented by $C^-$ is deducible from the information represented by $G$.

**Theorem 6** *A SG $G$ violates a negative constraint $C = (C', \rho)$ iff $\Phi(\mathcal{S}), \Phi(G) \vDash \Phi(C')$, where $C'$ is the SG underlying $C$ (and $\Phi(C')$ is the logical formula associated to this SG).*

Consistency relative to positive constraints can be explained with FOL, translating "projection" into a notion of "logical substitution" (Chein & Mugnier, 1992) between the formulas associated to graphs. We call an *S-substitution* $\sigma$ from $\Phi(G)$ into $\Phi(H)$ a substitution $\sigma$ of terms of $\Phi(G)$ by terms of $\Phi(H)$ such that constants of $\Phi(G)$ are kept invariant and, for any atom $t(e_1, ..., e_k)$ of $\Phi(G)$, there is $t' \le t$ such that $t'(\sigma(e_1), ..., \sigma(e_k))$ is an atom of $\Phi(H)$. The following property holds:

**Property 4** *Every projection $\pi$ from $G$ to $H$ defines an S-substitution $\sigma$ from $\Phi(G)$ to $\Phi(H)$. Assuming that $H$ is in normal form, the converse also holds.*

*Proof:* Let $\pi$ be a projection from $G$ to $H$. For each variable $x$ of $\Phi(G)$, let $c$ be the unique generic concept node such that $x = \Phi(c)$, then $\sigma(x) = \Phi(\pi(c))$. Reciprocally, provided that $H$ is in normal form, the application from concept nodes of $G$ to concept nodes of $H$, mapping each $c$ to the node $c'$ such that $\sigma(\Phi(c)) = \Phi(c')$ is a projection from $G$ to $H$. Note that, unless $H$ is in normal form, $c'$ is not uniquely defined when $\Phi(c')$ is a constant. $\square$

**Corollary 1** *A graph $G$ $\pi$-violates a constraint $C$ iff the S-substitution $\sigma$ from $\Phi(C_{(0)})$ into $\Phi(G)$ associated with $\pi$ cannot be extended to an S-substitution from $\Phi(C)$ into $\Phi(G)$.*

Another bridge can be built using rules. Indeed, a graph $G$ satisfies a positive constraint $C$ if and only if, considering $C$ as a rule, all applications of $C$ on $G$ produce a graph equivalent to $G$. Or, more specifically:

**Property 5** *A SG $G$ $\pi$-violates a positive constraint $C$ iff, considering $C$ as a rule, the application of $C$ on $G$ according to $\pi$ produces a graph not equivalent to $G$.*

*Proof:* Let $C$ be a constraint and $G$ be a SG such that $G$ satisfies $C$. If $\pi_0$ is a projection from $C_{(0)}$ into $G$, let us consider the graph $G'$ obtained by the application of $C$ (considered now as a rule) on $G$ according to $\pi_0$. Let us now build the following projection $\pi'$ from $G'$ into $G$: for each node $v$, $\pi'(v) = v$ if $v$ belongs to $G$; otherwise, $v$ is a copy of a node $w$ of $C_{(1)}$, and if $\pi$ is one of the projections from $C$ into $G$ that extends $\pi_0$, we have $\pi'(v) = \pi(w)$. Then $\pi'$ is a projection of $G'$ into $G$, and since $G$ trivially projects into $G'$, they are thus equivalent.

This proves the $\Leftarrow$ part of property 5. For the $\Rightarrow$ part, we use the following property, proved by Cogis and Guinaldo (1995). In their property (prop. 6 of their paper) the SGs considered are connected graphs, but the proof holds for non connected graphs.

**Property 6 (Cogis & Guinaldo, 1995)** *Let $G$ be a SG and $irr(G)$ be one of its equivalent irredundant subgraphs (if $G$ is not redundant then $irr(G) = G$). Then there exists a folding from $G$ to $irr(G)$, i.e. a projection $f$ from $G$ into $irr(G)$, such that the restriction of $f$ to nodes of $irr(G)$ is the identity (for every node $x$ of $irr(G)$, $f(x) = x$).*

Suppose now $G$ $\pi$-violates $C$. Since constraint violation is defined with respect to the irredundant form of a graph, we can consider, without loss of generality, that $G$ is irredundant. We denote by $G'$ the graph obtained by the application of $C$ (again, considered now as a rule) on $G$ according to $\pi$. We prove that "$G'$ equivalent to $G$" leads to a contradiction.

If $G'$ is equivalent to $G$, then there exists a projection from $G'$ into $G$. And since $G$ is an irredundant subgraph of $G'$, there exists a *folding* $f$ from $G'$ into $G$ (property 6). Consider now $\pi'$ the projection from $C$ to $G$ defined as follows: for any node $x$ of $C_{(0)}$, $\pi'(x) = f(\pi(x))$, otherwise let $x'$ be the copy of $x$ in $G'$, we have $\pi'(x) = f(x')$. Since for all $x$ in $C_{(0)}$, $f(\pi(x)) = \pi(x)$, $\pi'$ extends $\pi$. This contradicts the hypothesis "$G$ $\pi$-violates $C$". Thus $G'$ is not equivalent to $G$. $\qquad\square$

**Property 7** *If a SG $G$ satisfies a positive constraint $C$, then any graph in a $\{C\}$-derivation of $G$ is equivalent to $G$.*

*Proof:* Let $G = G_0$ ,..., $G_k$ be a $\{C\}$-derivation of $G$. From property 5, each $G_i$, $1 \le i \le k$, is equivalent to $G_{i-1}$, thus by transitivity, is equivalent to $G$. $\qquad\square$

Using soundness and completeness of the $\mathcal{SR}$ deduction, and properties 5 and 7, one obtains the following relation with FOL deduction.

**Theorem 7** *A SG $G$ violates a positive constraint $C$ iff there exists a SG $G'$ such that $\Phi(\mathcal{S}), \Phi(G), \Phi(C) \vDash \Phi(G')$ and not $\Phi(\mathcal{S}), \Phi(G) \vDash \Phi(G')$, where $\Phi(C)$ is the translation of $C$ considered as a rule.*

This theorem can be reformulated in terms of abductive inference (using in fact indirect abduction, see, for example, Konolige, 1996). Indeed, given a background theory $\Sigma = \Phi(\mathcal{S}), \Phi(G)$ and an observation $O = \neg\Phi(C)$, $G$ violates $C$ iff there is an abductive explanation for $O$ of the form $\neg F$, where $F$ is a formula belonging to $\text{FOL}(\wedge, \exists)$.

### 5.3 Computational Complexity

The problem "does a given graph satisfy a given constraint?" is co-NP-complete if this constraint is negative (since we must check the absence of projection), but becomes $\Pi_2^P$-complete for a positive one ($\Pi_2^P$ is co-NP$^{\text{NP}}$).

**Theorem 8 (Complexity in $\mathcal{SGC}$)** *$\mathcal{SGC}$-CONSISTENCY is $\Pi_2^P$-complete (but is co-NP-complete if all constraints are negative).*

*Proof:* Without change of complexity, one can consider that $\mathcal{C}$ is composed of only one positive constraint, say $C$. First recall that deciding whether a SG $G$ satisfies $C$ is done on the irredundant form of $G$. We shall consider two ways of integrating this fact in the complexity of $\mathcal{SGC}$-CONSISTENCY. One way is to assume that the irredundant form of $G$ is computed before the consistency check. This can be achieved with a number of calls to a projection oracle linear in the size of $G$ (Mugnier, 1995). But, since we have then to solve a function problem (compute the irredundant form of $G$) instead of a decision problem (is $G$ irredundant?), we prefer to integrate irredundancy into the consistency check: then, for a projection $\pi_0$ from the trigger of $C$ into $G$, the projection from $C$ to $G$ we look for does not necessarily extends $\pi_0$, but extends the composition of a projection from $G$ into one of its subgraphs (possibly equal to $G$ itself) and $\pi_0$.

First, $\mathcal{SGC}$-CONSISTENCY belongs to $\Pi_2^P$ since it corresponds to the language $L = \{x \mid \forall y_1 \; \exists y_2 \; R(x, \; y_1, \; y_2)\}$, where $x$ encodes an instance $(G, C)$ of the problem and $(x, \; y_1, \; y_2) \in R$ iff $y_1$ encodes a projection $\pi_0$ from $C_{(0)}$ into $G$ and $y_2$ encodes a projection $\pi_G$ from $G$ into one of its subgraphs and a projection $\pi$ from $C$ into $G$ s.t. $\pi[C_{(0)}] = \pi_G \circ \pi_0$. Note that if $G$ is in irredundant form, then $\pi_G$ is an automorphism.

Now, let us consider the problem $B_2^c$: given a boolean formula $E$, and a partition $\{X_1, \; X_2\}$ of its variables, is it true that for any truth assignment for the variables in $X_1$ there exists a truth assignment for the variables in $X_2$ s.t. $E$ is true? This problem is $\Pi_2^P$-complete, since its complementary $B_2$ is shown to be $\Sigma_2^P$-complete by Stockmeyer (1977). In order to build a polynomial reduction to $\mathcal{SGC}$-CONSISTENCY, we use a restriction of this problem to $k$-CNFs, i.e. conjunctions of disjunctions with at most $k$ literals per clause. Let us call 3-$SAT_2^c$ the special case where $E$ is a 3-CNF, in other words an instance of 3-SAT. Then 3-$SAT_2^c$ is also $\Pi_2^P$-complete. Indeed, in the same paper (Th. 4.1), Stockmeyer shows that $B_2$ with $E$ restricted to a 3-disjunctive normal form (3-DNF) remains $\Sigma_2^P$-complete. Since the negation of a 3-DNF is a 3-CNF, it follows that the complementary problem $B_2^c$ with $E$ restricted to a 3-CNF is $\Pi_2^P$-complete.

Let us now reduce 3-$SAT_2^c$ to $\mathcal{SGC}$-CONSISTENCY. The transformation used is very similar to the one from 3-SAT to $\mathcal{SG}$-DEDUCTION (proof of theorem 2), illustrated in Fig. 4. Let $E$ be an instance of 3-SAT. Let $G(E)$ and $Q(E)$ be the SGs obtained by the transformation described in the proof of Th. 2. The constraint $C(E) = (Q(E), \rho(X_2))$ is obtained by adding a coloration to $Q(E)$: all relation nodes obtained from clauses (nodes typed $\mathtt{C}_i$) and all nodes obtained from variables in $X_2$ (concept nodes typed x or xv and relation nodes typed val) are colored by 1 (i.e. belong to the obligation). Once again, having clauses of bounded size leads to a polynomial transformation. The simple graph $G$ and the positive constraint $C$ presented in Fig. 12 are obtained from the 3-SAT formula $(a \vee b \vee \neg c) \wedge (\neg a \vee c \vee \neg d)$ and the partition $X_1 = \{a, b\}, X_2 = \{c, d\}$.

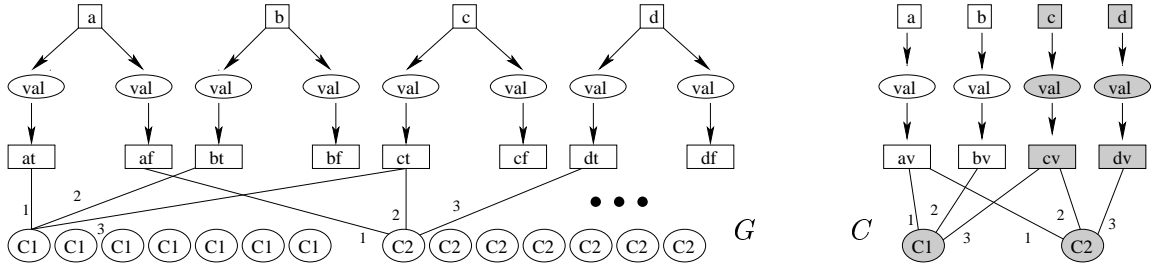

Figure 12: Example of transformation from $3\text{-}SAT_2^c$ to $\mathcal{SGC}$-consistency

Each truth assignment of the variables of $E$ s.t. $E$ is true naturally gives a projection from $C$ into $G$, and reciprocally (as indicated in the proof of Th. 2). Furthermore, any truth assignment for the variables of $X_1$ naturally gives a projection from $C_{(0)}$ into $G$, and reciprocally. Thus, the question "is it true that for any truth assignment for the variables in $X_1$ there exists a truth assignment for the variables in $X_2$ s.t. $E$ is true?" is equivalent to the question "is it true that for any projection $\pi_0$ from $C_{(0)}$ into $G$ there exists a projection from $C$ into $G$ extending $\pi_0$?". □

Note that this reduction is less straightforward than the one we proposed in (Baget & Mugnier, 2001), but it will be used as a basis for the proof of Th. 12.

**Corollary 2** *Deduction in $\mathcal{SGC}$ is $\Pi_2^P$-complete.*

## 6. Rules and Constraints: $\mathcal{SEC}/\mathcal{SRC}/\mathcal{SREC}$

In presence of constraints, the two kinds of rules, inference rules $\mathcal{R}$ and evolution rules $\mathcal{E}$, define two alternative models.

### 6.1 Definitions and Notations

In $\mathcal{SEC}$, $\mathcal{G}$ is seen as the initial world, root of a potentially infinite tree of possible worlds, and $\mathcal{E}$ describes the possible evolutions from one world to others. The deduction problem asks whether there is *a path of consistent worlds from $\mathcal{G}$ to a world satisfying $Q$*.

**Definition 12 ($\mathcal{SEC}$-deduction)** *Let $\mathcal{K} = (\mathcal{G}, \mathcal{E}, \mathcal{C})$ be a KB, and let $Q$ be a SG. $Q$ can be deduced from $\mathcal{K}$ if there is an $\mathcal{E}$-derivation $\mathcal{G} = G_0, \ldots, G_k$ such that, for $0 \leq i \leq k$, $(G_i, \mathcal{C})$ is consistent and $Q$ can be deduced from $G_k$.*

In $\mathcal{SRC}$, $\mathcal{G}$ provided with $\mathcal{R}$ is a finite description of a potentially infinite world, that has to be consistent. Applying a rule to $\mathcal{G}$ can create inconsistency, but a further application of a rule may restore consistency. Let us formalize this notion of *consistency restoration*.

Suppose there is a $\pi$-violation of a positive constraint $C$ in $\mathcal{G}$; this violation $(C, \pi)$ is said to be $\mathcal{R}$-*restorable* if there exist an $\mathcal{R}$-derivation from $\mathcal{G}$ into a SG $H$ and a projection $\pi'$ from $H$ into $\mathrm{irr}(H)$ such that the projection $\pi' \circ \pi$ of the trigger of $C$ into $\mathrm{irr}(H)$ can be extended to a projection of $C$ as a whole. The violation of a negative constraint can never be restored. Note that the $\mathcal{R}$-restoration can create new violations, that must themselves be proven $\mathcal{R}$-restorable.

**Definition 13 ($\mathcal{SRC}$-CONSISTENCY and $\mathcal{SRC}$-DEDUCTION)** *A KB $\mathcal{K} = (\mathcal{G}, \mathcal{R}, \mathcal{C})$ is consistent if, for any SG $H$ that can be $\mathcal{R}$-derived from $\mathcal{G}$, for every constraint $C \in \mathcal{C}$, for every $\pi$-violation of $C$ in $H$, $(C, \pi)$ is $\mathcal{R}$-restorable. A SG $Q$ can be deduced from $\mathcal{K}$ if $\mathcal{K}$ is consistent and $Q$ can be deduced from $(\mathcal{G}, \mathcal{R})$.*

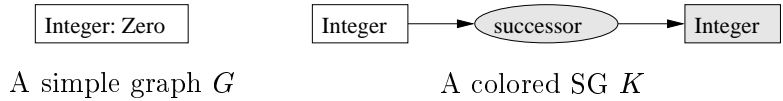

A simple graph $G$      A colored SG $K$

Figure 13: Consistency in $\mathcal{SEC}/\mathcal{SRC}$

Consider for instance a KB containing the SG $G$ in Fig. 13, expressing the existence of the number 0, a constraint and a rule, both represented by the colored SG $K$. The constraint asserts that *for every integer $n$, there must be an integer $n'$, successor of $n$*. If the rule is an evolution rule, $G$ is seen as an inconsistent initial world (there is no successor of 0 in $G$) and nothing will be deduced from this KB. If the rule is an inference rule, its application immediately repairs the constraint violation, while creating a new integer, that has no successor, thus a new violation. Finally, every constraint violation could eventually be repaired by a rule application, and the KB should be proven consistent.

Let us point out that the $\mathcal{SR}$ model is obtained from $\mathcal{SRC}$ or $\mathcal{SEC}$ when $\mathcal{C}$ is empty, and $\mathcal{SGC}$ is obtained from $\mathcal{SRC}$ (resp. $\mathcal{SEC}$) when $\mathcal{R}$ (resp. $\mathcal{E}$) is empty.

The $\mathcal{SREC}$ model combines both derivation schemes of the $\mathcal{SRC}$ and $\mathcal{SEC}$ models. Now, $\mathcal{G}$ describes an initial world, inference rules of $\mathcal{R}$ complete the description of any world, constraints of $\mathcal{C}$ evaluate the consistency of a world, evolution rules of $\mathcal{E}$ try to make a consistent world evolve into a new, consistent one. The deduction problem asks whether $\mathcal{G}$ can evolve into a consistent world satisfying the goal.

**Definition 14 ($\mathcal{SREC}$-DEDUCTION)** *A SG $G'$ is an immediate $\mathcal{RE}$-evolution from a SG $G$ if there exists an $\mathcal{R}$-derivation from $G$ into $G''$ and an immediate $\mathcal{E}$-derivation from $G''$ into $G'$. An $\mathcal{RE}$-evolution from a SG $G$ to a SG $G'$ is a sequence of SGs $G = G_0, \ldots, G_k = G'$ such that, for $0 \le i \le k$, $(G_i, \mathcal{R}, \mathcal{C})$ is consistent and, for $1 \le i \le k$, $G_i$ is an immediate $\mathcal{RE}$-evolution from $G_{i-1}$. Given a KB $\mathcal{K} = (\mathcal{G}, \mathcal{R}, \mathcal{E}, \mathcal{C})$, a SG $Q$ can be deduced from $\mathcal{K}$ if there is an $\mathcal{RE}$-evolution $\mathcal{G} = G_0, \ldots, G_k$ such that $Q$ can be deduced from $(G_k, \mathcal{R})$.*

When $\mathcal{E} = \emptyset$ (resp. $\mathcal{R} = \emptyset$), one obtains the $\mathcal{SRC}$ model (resp. $\mathcal{SEC}$ ).

## 6.2 Consistency in $\mathcal{SRC}$ and Logics

To translate $\mathcal{SRC}$-DEDUCTION in logics, a starting point could be to extend the logical translation of $\mathcal{SGC}$-CONSISTENCY given in Th. 7 to a translation of $\mathcal{SRC}$-CONSISTENCY. However, the following theorem points out the limitations of this approach.

**Theorem 9** *Let $\mathcal{K} = (\mathcal{G}, \mathcal{R}, \mathcal{C})$ be a KB. If there exists a SG $G'$ such that $\Phi(\mathcal{S}), \Phi(\mathcal{G}), \Phi(\mathcal{R}), \Phi(\mathcal{C}) \models \Phi(G')$ and not $\Phi(\mathcal{S}), \Phi(\mathcal{G}), \Phi(\mathcal{R}) \models \Phi(G')$, where $\Phi(\mathcal{C})$ is the translation of the constraints of $\mathcal{C}$ considered as rules, then $\mathcal{K}$ is inconsistent. However, the converse is false in the general case.*

*Proof:* We first prove the positive part of this theorem. If there exists such a graph $G'$, then (Th. 4) there is a $(\mathcal{R} \cup \mathcal{C})$-derivation (considering the colored graphs of $\mathcal{C}$ as rules) $\mathcal{G} = G_0, \ldots, G_k$ such that $G'$ projects to $G_k$. See that $G_k$ cannot be deduced from $(\mathcal{G}, \mathcal{R})$, otherwise $G'$ would also be deducible from $(\mathcal{G}, \mathcal{R})$. Let us consider the first $G_i$ from this derivation that is not deducible from $(\mathcal{G}, \mathcal{R})$. Then $G_i$ is obtained from $G_{i-1}$ (a graph $\mathcal{R}$-deducible from $\mathcal{G}$) by applying a rule $C_q \in \mathcal{C}$ following a projection $\pi$. Since $G_{i-1}$ is deducible from $(\mathcal{G}, \mathcal{R})$, then there exists a graph $H$ $\mathcal{R}$-derivable from $\mathcal{G}$ such that $G_{i-1}$ projects into $H$. Let us call $\pi'$ such a projection, and consider the projection $\pi'' = \pi' \circ \pi$ of the hypothesis/trigger of the rule/constraint $C_q$ into $H$. We now have to prove that 1) $H$ $\pi''$-violates $C_q$, and 2) this violation is not $\mathcal{R}$-restorable. Suppose 2) is false. Then there would exist a graph $H'$ $\mathcal{R}$-derived from $H$ such that $\pi''$ can be extended to a projection $\Pi$ of $C_q$ as a whole in the irredundant form of $H'$. This is absurd, since $\Pi$ is a projection of $G_i$ in a graph $\mathcal{R}$-derivable from $\mathcal{G}$.

The counterexample presented in Fig. 14 is sufficient to prove the negative part of the theorem.

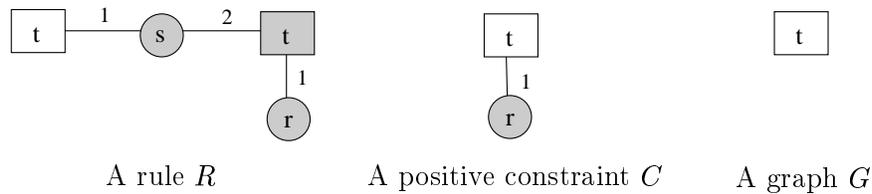

A rule $R$      A positive constraint $C$      A graph $G$

Figure 14: A counterexample to Th. 9

It is immediate to check that every graph that can be $\{R, C\}$-derived from $G$ can also be $\{R\}$-derived from $G$. However, the projection of the trigger of $C$ into the unique node of $G$ defines a violation of $C$ that will never be restored. □

We will study in the next section (Th. 11) a particular case of rules where the converse of Th. 9 is true.

## 6.3 Undecidability of the Associated Deduction Problems

**Theorem 10 (Complexity in $\mathcal{SEC}/\mathcal{SRC}$)** *$\mathcal{SEC}$-DEDUCTION is semi-decidable. Both $\mathcal{SRC}$-CONSISTENCY and $\mathcal{SRC}$-DEDUCTION are truly undecidable.*

*Proof:* $\mathcal{SEC}$ includes $\mathcal{SR}$ thus $\mathcal{SEC}$-deduction is not decidable. When $Q$ is deducible from $\mathcal{K}$, a breadth-first search of the tree of all derivations from $\mathcal{K}$, each graph being checked for consistency, ensures that $G_k$ is found in finite time. For $\mathcal{SRC}$, we show that checking consistency is truly undecidable. Let $\mathcal{K}$ be a KB where $\mathcal{C}$ contains a positive constraint $C^+$ and a negative constraint $C^-$, both with an empty trigger. For proving consistency, one has to prove that $C^- \not\succeq (\mathcal{G}, \mathcal{R})$, and the algorithm does not necessarily stop in this case (from semi-decidability of deduction in $\mathcal{SR}$). The same holds for the complementary problem (proving inconsistency) taking $C^+$ instead of $C^-$, hence the undecidability. □

As a generalization of $\mathcal{SRC}$, deduction in $\mathcal{SREC}$ is truly undecidable. Next section studies a decidable fragment of $\mathcal{SREC}$, which in particular was sufficient for the Sysiphus-I modelization.

## 7. Decidability and Complexity of some Particular Cases

A rule application may add redundant information to a graph. In general, detecting redundancy is difficult (recall determining whether a graph is redundant is an NP-complete problem), but there are some trivial cases, which we will get rid of, since they may create artificially infinite derivations. First, once a rule has been applied to a graph according to a given projection, it can be applied again to the resulting graph, according to the same projection, and this indefinitely. These further applications obviously produce redundant information. They are said to be *useless*. Another case of trivial redundancy in a graph is that of *twin* relation nodes, i.e. with exactly the same neighbors in the same order. Consider for instance a rule of kind "if $r(x,\ y)$ then $r(x,\ y)$". This rule can be applied indefinitely, even if useless applications are avoided, but all applications create twin relation nodes. In what follows, we consider that the construction of the graph resulting from a rule application prevents the generation of twin relation nodes, and that a derivation does not comprise any useless rule application.

Given a set of rules $\mathcal{R}$ and an $\mathcal{R}$-derivation leading to a SG $H$, $H$ is said to be *closed* if no rules of $\mathcal{R}$ can be applied to $H$ in an original way, i.e. all applications of any rule of $\mathcal{R}$ on $H$ are useless. More formally, $H$ is closed w.r.t. $\mathcal{R}$ and w.r.t. an $\mathcal{R}$-derivation $H_0\ \pi_1\ ...\ \pi_k H_k = H$, where $H_i\ (1 \le i \le k)$ is the graph obtained by the application of a rule of $\mathcal{R}$ on $H_{i-1}$ according to the projection $\pi_i$, if for every rule $R$ of $\mathcal{R}$, for every projection $\pi$ from $R_{(0)}$ into $H$, there exists a projection $\pi_i$ from $R_{(0)}$ to $H_{i-1}$ $(1 \le i \le k)$, such that $\pi = \pi_i$.

Given a set of rules $\mathcal{R}$ and a graph $G$, if a closed graph is $\mathcal{R}$-derivable from $G$, then it is unique. Moreover, if this graph is derivable with $n$ rule applications, then $n$ is the maximal length of an $\mathcal{R}$-derivation, and all derivations of length $n$ lead to it. When it exists, we call it the *closure* of $G$ w.r.t. $\mathcal{R}$, which we note $G_{\mathcal{R}}^*$.

Let us also define another notion, related to the fact that we are interested in irredundant graphs. In this perspective, let us say that an irredundant graph $H$ is *full* w.r.t. a set of rules $\mathcal{R}$ if every graph that can be obtained by applying one of those rules on $H$ is equivalent to $H$. Assuming that $G$ is an irredundant graph and that graphs obtained by a rule application are put into irredundant form, if a full graph can be derived from $G$ then it is unique.

Informally, the notion of a closed graph translates the fact that nothing can be added that has not been already added, whereas the notion of a full graph says that nothing can be added that really adds new information to the graph. When the closure of a graph $G$ exists, then the irredundant form of this closure is exactly the full graph derivable from $G$. But note that when the full graph exists, the closure does not necessarily exists (see proof of Prop. 10).

## 7.1 Finite Expansion Sets

The notion of a full graph being more general than the notion of a closure, we can generalize the definition of finite expansion sets used in a previous paper (Baget & Mugnier, 2001), and adopt the following one:

**Definition 15 (Finite expansion sets)** *A set of rules $\mathcal{R}$ is called a* finite expansion set *if, for every SG $G$, there exists an $\mathcal{R}$-derivation $G \ldots G'$ such that $irr(G')$ is full w.r.t. $\mathcal{R}$. We denote by $G^{\mathcal{R}}$ this full graph.*

If $\mathcal{R}$ is a finite expansion set (f.e.s), deduction in $\mathcal{SR}$ becomes decidable (but it is not a *necessary* condition for decidability). Indeed, in order to determine whether a SG $Q$ is deducible from a KB $(\mathcal{G}, \mathcal{R})$, it suffices to compute $\mathcal{G}^{\mathcal{R}}$, then to check the existence of a projection from $Q$ to $\mathcal{G}^{\mathcal{R}}$. Similarly, consistency checks in $\mathcal{SRC}$ are done on $\mathcal{G}^{\mathcal{R}}$.

**Property 8 (Finite expansion sets)** *Let $\mathcal{K} = (\mathcal{G}, \mathcal{R}, \mathcal{C})$ be a KB where $\mathcal{R}$ is a finite expansion set. Then $\mathcal{K}$ is consistent iff $(\{\mathcal{G}^{\mathcal{R}}\}, \mathcal{C})$ is consistent, and a SG $Q$ can be deduced from $(\mathcal{G}, \mathcal{R})$ iff $Q$ can be deduced from $(\{\mathcal{G}^{\mathcal{R}}\})$.*

This property allows us to prove that the converse of Th. 9 is true when $\mathcal{R}$ is restricted to a finite expansion set.

**Theorem 11** *Let $\mathcal{K} = (\mathcal{G}, \mathcal{R}, \mathcal{C})$ be a KB, where $\mathcal{R}$ is a finite expansion set. Then $\mathcal{K}$ is inconsistent iff there exists a SG $G'$ such that $\Phi(\mathcal{S}), \Phi(\mathcal{G}), \Phi(\mathcal{R}), \Phi(\mathcal{C}) \vDash \Phi(G')$ and not $\Phi(\mathcal{S}), \Phi(\mathcal{G}), \Phi(\mathcal{R}) \vDash \Phi(G')$, where $\Phi(\mathcal{C})$ is the translation of the constraints of $\mathcal{C}$ considered as rules.*

*Proof:* ($\Leftarrow$) holds as a particular case of Th. 9. Let us now prove the ($\Rightarrow$) part. Since $\mathcal{K}$ is inconsistent, the previous property asserts that $(\{\mathcal{G}^{\mathcal{R}}\}, \mathcal{C})$ is inconsistent. Th. 7 ensures that there exists a graph $H$ such that 1) $\Phi(\mathcal{S}), \Phi(\mathcal{G}^{\mathcal{R}}), \Phi(\mathcal{C}) \vDash \Phi(H)$, and 2) $\Phi(\mathcal{S}), \Phi(\mathcal{G}^{\mathcal{R}}) \nvDash \Phi(H)$. Since $\Phi(\mathcal{S}), \Phi(\mathcal{G}), \Phi(\mathcal{R}) \vDash \Phi(\mathcal{G}^{\mathcal{R}})$ (Th. 4), we obtain $\Phi(\mathcal{S}), \Phi(\mathcal{G}), \Phi(\mathcal{R}), \Phi(\mathcal{C}) \vDash \Phi(H)$. Let us now suppose that $\Phi(\mathcal{S}), \Phi(\mathcal{G}), \Phi(\mathcal{R}) \vDash \Phi(H)$, and prove that it is absurd. In that case, there would be a graph $G'$ $\mathcal{R}$-derivable from $\mathcal{G}$ such that $H$ projects into $G'$ (Th. 4 again). And since $H \geq G' \geq \mathcal{G}^{\mathcal{R}}$, we should have $\Phi(\mathcal{S}), \Phi(\mathcal{G}^{\mathcal{R}}) \vDash \Phi(H)$ (Th. 1): this is absurd. $\quad\square$

More generally, one obtains the following decidability results, depending on whether $\mathcal{R}$, $\mathcal{E}$, or $\mathcal{R} \cup \mathcal{E}$ is a finite expansion set.

## Property 9 (Complexity with finite expansion rule sets)

- *When $\mathcal{R}$ is a f.e.s, deduction in $\mathcal{SR}$ is decidable, consistency and deduction in $\mathcal{SRC}$ are decidable, deduction in $\mathcal{SREC}$ is semi-decidable.*

- *When $\mathcal{E}$ is a f.e.s, deduction in $\mathcal{SEC}$ is decidable, but remains truly undecidable in $\mathcal{SREC}$.*

- *When $\mathcal{R} \cup \mathcal{E}$ is a f.e.s, deduction in $\mathcal{SREC}$ is decidable.*

*Proof:* Suppose $\mathcal{R}$ is a f.e.s. Decidability of problems in $\mathcal{SR}$ and $\mathcal{SRC}$ follows from property 8. In $\mathcal{SREC}$, when the answer is "yes", it can be obtained in finite time; we proceed as for $\mathcal{SEC}$ (see proof of theorem 10) but consistency checks are done on the full graph instead of the graph itself.

Now, suppose $\mathcal{E}$ is a f.e.s. $\mathcal{G}^{\mathcal{E}}$ exists, thus the derivation tree in $\mathcal{SEC}$ is finite, and consistency checks may only cut some parts of this tree. Deduction in $\mathcal{SREC}$ remains undecidable because when $\mathcal{E} = \emptyset$, one obtains the $\mathcal{SRC}$ model, in which deduction is truly undecidable.

Finally, if $\mathcal{R} \cup \mathcal{E}$ is a f.e.s, $\mathcal{G}^{\mathcal{R} \cup \mathcal{E}}$ exists, thus the derivation tree is finite, and consistency checks may only cut parts of this tree. $\square$

Note that the condition "$\mathcal{R} \cup \mathcal{E}$ is a finite expansion set" is stronger than "both $\mathcal{R}$ and $\mathcal{E}$ are finite expansion sets". The following property justifies this condition:

**Property 10** *If both $\mathcal{R}$ and $\mathcal{E}$ are finite expansion sets, then $\mathcal{SREC}$-DEDUCTION is not necessarily decidable.*

*Proof:* We build a reduction from WORD PROBLEM IN A SEMI-THUE SYSTEM (Thue, 1914) to $\mathcal{SREC}$-DEDUCTION, where the obtained rule sets $\mathcal{R}$ and $\mathcal{E}$ are both finite expansion sets. This reduction relies on the one built for proving the semi-decidability of $\mathcal{SR}$-DEDUCTION (theorem 5).

Let us first present the two kind of finite expansion sets used in this reduction. $\mathcal{E}$ is a finite expansion set since only relation nodes are present in the conclusion of rules: $\mathcal{E}$ is indeed a particular case of range-restricted rules (see Prop. 11). $\mathcal{R}$ is also a finite expansion set since, for every rule in $\mathcal{R}$, the hypothesis is disconnected from the conclusion (we call these rules *disconnected*). Note this time that, though $\mathcal{R}$ is trivially a f.e.s., the closure of a graph w.r.t. $\mathcal{R}$ does not necessarily exist.

Recall the WORD PROBLEM takes as input two words $m$ and $m'$ and a set of rules $\Gamma = \{\gamma_1, \ldots, \gamma_k\}$, each rule $\gamma_i$ being a pair of words $(\alpha_i, \beta_i)$, and asks whether there is a derivation from $m$ to $m'$. There is an immediate derivation from $m$ to $m'$ (we note $m \to m'$) if, for some $\gamma_j$, $m = m_1 \alpha_j m_2$ and $m' = m_1 \beta_j m_2$. A *derivation* from $m$ to $m'$ (we note $m \leadsto m'$) is a sequence $m = m_0 \to m_1 \to \ldots \to m_p = m'$.

We have shown how this problem can be expressed in the $\mathcal{SR}$ model: to a word $m = x_1 \ldots x_k$ is associated the graph $\mathcal{G}(m)$, and to any rule $\gamma = (y_1 \ldots y_p, \ z_1 \ldots z_q)$ is associated the graph rule $\mathcal{U}(\gamma)$, as represented in Fig. 15 (where $\top$ is greater than all other concept types). Then $m \leadsto m'$ iff $\mathcal{G}(m') \geq (\mathcal{G}(m), \mathcal{U}(\Gamma))$ (see proof of Th. 5).

Let us now split each obtained rule $\mathcal{U}(\gamma)$ into one disconnected inference rule $\mathcal{R}(\gamma)$ and one range-restricted evolution rule $\mathcal{E}(\gamma)$. We distinguish in the hypothesis of $\mathcal{E}(\gamma)$ two subgraphs: the *origin*, which corresponds to the hypothesis of $\mathcal{U}(\gamma)$, and the *destination*, which corresponds to the conclusion of $\mathcal{R}(\gamma)$. It is easy to check that one part of the above equivalence still holds: $m \leadsto m' \Rightarrow \mathcal{G}(m') \geq (\mathcal{G}(m), \mathcal{R}(\Gamma) \cup \mathcal{E}(\Gamma))$. However, the converse is no longer valid: check by exemple that, if $\Gamma = \{\gamma = (a, c)\}$, we have $\mathcal{G}(c) \geq (\mathcal{G}(aba), \mathcal{R}(\Gamma) \cup \mathcal{E}(\Gamma))$ though $aba \not\leadsto c$ (apply $\mathcal{R}(\gamma)$ once, then two times $\mathcal{E}(\gamma)$ using different nodes [a] but the same node [c]).

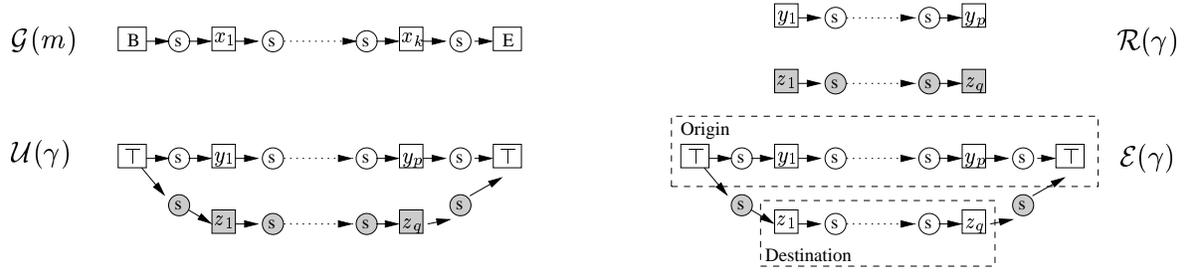

Figure 15: Transformations from the word problem into models of the $\mathcal{SG}$ family

We thus need the notion of a *good* application of a rule $\mathcal{E}(\gamma)$: it is such that the projection of its destination part is a mapping to a subgraph that was obtained by applying the rule $\mathcal{R}(\gamma)$, and that was never used to project the destination part of any rule of $\mathcal{E}(\Gamma)$, including $\mathcal{E}(\gamma)$ itself. Moreover, the origin and destination must be projected into disjoint paths (i.e. a node of the origin cannot have the same image has a node of the destination). If we restrict ourselves in some way to good applications of rules of $\mathcal{E}(\Gamma)$, then we can verify that $\mathcal{G}(m') \geq (\mathcal{G}(m), \mathcal{U}(\Gamma))$ iff $\mathcal{G}(m') \geq (\mathcal{G}(m), \mathcal{R}(\Gamma) \cup \mathcal{E}(\Gamma))$.

This restriction is obtained by using constraints, that will allow every good application of a rule of $\mathcal{E}(\Gamma)$, and be violated by the obtained graph otherwise. Let us note $\mathcal{R}'(\Gamma)$ and $\mathcal{E}'(\Gamma)$ the new sets of inference rules and evolution rules. The new transformation is described in Fig. 16. It allows to obtain the following result: $m \leadsto m' \Leftrightarrow \mathcal{G}'(m') \geq (\mathcal{G}'(m), \mathcal{R}'(\Gamma), \mathcal{E}'(\Gamma), \{C_+, C_-\})$. The names of relation types $=, \in$ and $\rightarrow$ have been chosen to give an intuitive idea of their role but they are just types as others. A relation node ($\in$) from a node [$z_j$] to a node [$\gamma_i$] means that the letter $z_j$ has been obtained by applying the rule $\gamma_i$. A relation node ($\rightarrow$) from [$y_k$] to [$\gamma_i$] means that the letter $y_k$ belongs to the subword on which the rule $\gamma_i$ has been applied. $=$ is used to indicate that two concept nodes have to be projected on the same node (in CG terms, we would see it as a co-reference link). The evolution rule $\mathcal{E}'(\gamma_i)$, starting from a path representing the subword $\alpha_i$ used to apply $\gamma_i$ and from the representation of $\beta_i$ generated by $\mathcal{R}'(\gamma_i)$, produces the two relation nodes typed $s$ simulating the application of $\mathcal{U}(\gamma_i)$, thus $\gamma_i$, and the relation nodes typed $\rightarrow$ which mark the representation of $\alpha_i$ as used by an application of $\gamma_i$. The negative constraint $C_-$ prevents an application of $\mathcal{E}'(\gamma_i)$ in which two nodes of the origin and destination parts have the same image (a node necessarily obtained by some application of the rule $\mathcal{R}'(\gamma_i)$); while the positive constraint $C_+$ prevents such a subgraph to be used twice for applying $\mathcal{E}'(\gamma_i)$ with different projections of its origin: it says that in this case, the two projections of the origins must be the same. □

## 7.2 Range Restricted Rules

Let us now focus on the rules that were used to solve the Sisyphus-I problem. A bicolored graph (rule or constraint) is said to be *range restricted* (r.r.) if its second part (conclusion or obligation) does not comprise any generic concept node. We use this expression by analogy with the so-called rules in Datalog, where all variables of the head must appear in the body (Abiteboul et al., 1995). Such rules are also called *safe* in the literature. Consider for instance the rules of Fig. 5: $R_1$ and $R_2$ are range restricted, while $R_3$ is not.

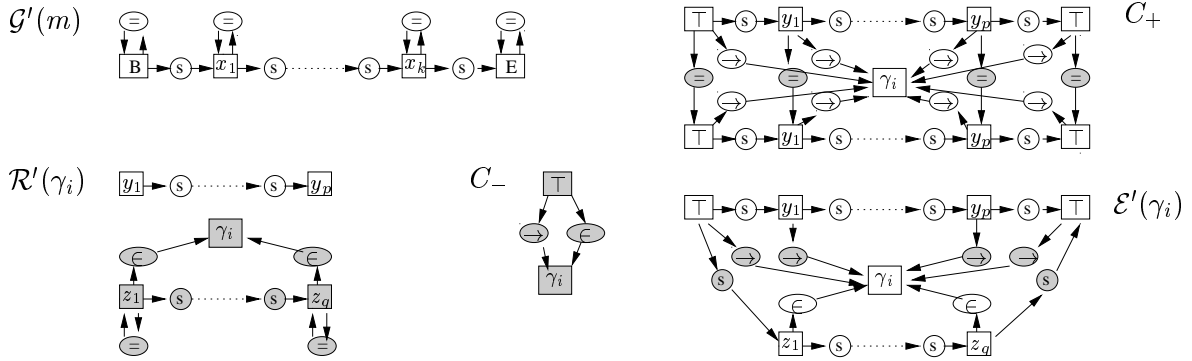

Figure 16: Reduction from the word problem to $\mathcal{SREC}$-deduction, with $\mathcal{E}$ and $\mathcal{R}$ f.e.s.

Also notice that a range restricted rule $R$ can be decomposed into an equivalent set of rules $\mathcal{D}(R)$ with exactly one node in conclusion (either an individual concept node or a relation node). There is one rule for each node of the conclusion of $R$: for each individual node $c$, one rule with same hypothesis as $R$ and a conclusion restricted to $c$; for each relation node $r$, one rule whose hypothesis is the disjoint union of the hypothesis of $R$ and of all individual concept nodes of the conclusion of $R$, and conclusion is $r$, with same neighbors as in $R$. The logical interpretation of such rules are (function free) range restricted Horn rules. If a SG $Q$ is deducible from a set of r.r. rules $\mathcal{R}$, then it is deducible from the set of their decompositions $\mathcal{D}(\mathcal{R})$, and reciprocally. However, as soon as constraints are involved, this equivalence does not hold any more.

**Property 11** *A set of range restricted rules is a finite expansion set.*

*Proof:* Since all graphs are put into normal form, an individual marker appears at most once in a graph. The number of individual nodes created by the set of rules is bounded by $M = |\mathcal{R}| \times max_{R \in \mathcal{R}}|R_{(1)}|$. So the number of relation nodes created (no twin relation nodes are created) is bounded by $N = \sum_{n=1}^{k} P_n(|V_C(G)| + M)^n$, where $P_n$ is the number of relation types with a given arity $n$ appearing in a rule conclusion, and $k$ is the greatest arity of such a relation type. So the closure of a graph can be obtained with a derivation of length $L \leq N + M$. We thus obtain $\mathcal{G}^{\mathcal{R}}$ in finite time. □

Note that, contrary to general finite expansion sets, existence of the closure and existence of the full graph are equivalent notions in the case of range restricted rules. It follows from the proof of property 11 that the length of a derivation from $\mathcal{G}$ to $\mathcal{G}^{\mathcal{R}}$ is in $\mathcal{O}(n^{k+1})$, where $n$ is the size of $(\mathcal{G}, \mathcal{R})$ and $k$ is the greatest arity of a relation type appearing in a rule conclusion. This rough upper bound could be refined but it is sufficient to obtain the following property, which will be used throughout the proofs of complexity results involving range restricted rules.

**Property 12** *Under the assumption that the maximum arity of relation types is a constant, given a range restricted set of rules $\mathcal{R}$, the length of an $\mathcal{R}$-derivation from $\mathcal{G}$ is polynomially related to the size of $(\mathcal{G}, \mathcal{R})$.*

In what follows, we assume that the arity of relation types is bounded by a constant.

**Theorem 12 (Complexity with range restricted rules)** *When $\mathcal{E}$ and $\mathcal{R}$ are range restricted rules:*

- *Deduction in $\mathcal{SR}$ is NP-complete.*

- *Consistency and deduction in $\mathcal{SRC}$ are $\Pi_2^P$-complete.*

- *Deduction in $\mathcal{SEC}$ and $\mathcal{SREC}$ is $\Sigma_3^P$-complete.*

*Proof:* The following results heavily rely on Prop. 12: all derivations involved being of polynomial length, they admit a polynomial certificate (the sequence of projections used to build the derivation).

*NP-completeness of $\mathcal{SR}$-deduction.* The problem belongs to NP. Indeed, a polynomial certificate is given by a derivation from $\mathcal{G}$ to a graph $G'$, followed by a projection from the goal to $G'$. When $\mathcal{R} = \emptyset$, one obtains $\mathcal{SG}$-deduction (projection checking), thus the NP-completeness.

*$\Pi_2^P$-completeness of $\mathcal{SRC}$-consistency and $\mathcal{SRC}$-deduction.* Recall the consistency check involves the irredundant form of $\mathcal{G}$. In order to lighten the problem formulation, we assume here that all SGs are irredundant, but irredundancy can be integrated without increasing the consistency check complexity: see the proof of theorem 8. $\mathcal{SRC}$-consistency belongs to $\Pi_2^P$ since it corresponds to the language $L = \{x \mid \forall y_1 \ \exists y_2 \ R(x, \ y_1, \ y_2)\}$, where $x$ encodes an instance $(\mathcal{G}; \mathcal{R}; \mathcal{C})$ of the problem and $(x, \ y_1, \ y_2) \in R$ iff $y_1 = (d_1; \pi_0)$, where $d_1$ is a derivation from $\mathcal{G}$ to $G'$, $\pi_0$ is a projection from the trigger of a constraint $C_{i(0)}$ into $G'$, $y_2 = (d_2; \pi_1)$, $d_2$ is a derivation from $G'$ to $G''$ and $\pi_1$ is a projection from $C_i$ into $G''$ s.t. $\pi_1[C_{i(0)}] = \pi_0$. $R$ is polynomially decidable and polynomially balanced (since the lengths of $d_1$ and $d_2$ are polynomial in the size of the input). When $\mathcal{R} = \emptyset$, one obtains the problem $\mathcal{SGC}$-consistency, thus the $\Pi_2^P$-completeness. Since $\mathcal{SRC}$-deduction consists in solving two independent problems, $\mathcal{SRC}$-consistency ($\Pi_2^P$-complete) and $\mathcal{SR}$-deduction (NP-complete), and since NP is included in $\Pi_2^P$, $\mathcal{SRC}$-deduction is also $\Pi_2^P$-complete.

*$\Sigma_3^P$-completeness of $\mathcal{SEC}$-deduction.* As for $\mathcal{SRC}$ (see above), we assume that all SGs are irredundant. The question is "are there a derivation from $\mathcal{G}$ to a SG $G'$ and a projection from $Q$ into $G'$, such that for all $G_i$ of this derivation, for all constraint $C_j$, for all projection $\pi$ from $C_{j(0)}$ into $G_i$, there exists a projection $\pi'$ from $C_j$ into $G_i$ s.t. $\pi'[C_{j(0)}] = \pi$?". $R$ is polynomially decidable and polynomially balanced (since the size of the derivation from $\mathcal{G}$ to $G'$ is polynomially related to the size of the input). Thus, $\mathcal{SEC}$-deduction is in $\Sigma_3^P$. In order to prove the completeness, we build a reduction from a special case of the problem $B_3$, where the formula is a 3-CNF (i.e. an instance of 3-SAT): given a formula $E$, which is a conjunction of clauses with at most 3 literals, and a partition $\{X_1, X_2, X_3\}$ of its variables, does there exist a truth assignment for the variables in $X_1$, such that for all truth assignment for the variables of $X_2$, there exists a truth assignment for the variables of $X_3$ such that $E$ is true? This problem is $\Sigma_3^P$-complete (Stockmeyer, 1977, theorem 4.1). Let us call it 3-SAT$_3$.

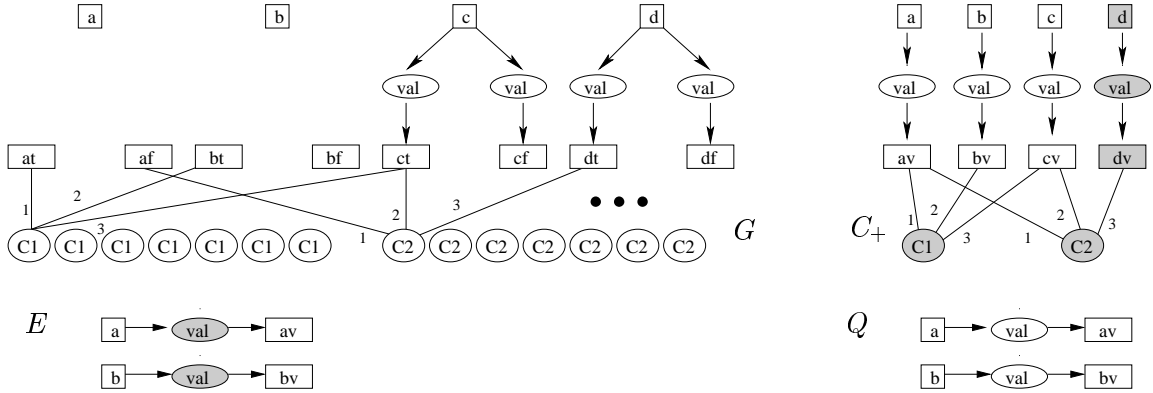

Figure 17: Example of transformation from 3-$SAT_3$ to $\mathcal{SEC}$-deduction

The transformation we use is illustrated in Fig. 17. The 3-SAT formula used is again $(a \vee b \vee \neg c) \wedge (\neg a \vee c \vee \neg d)$, and the partition is $X_1 = \{a, b\}, X_2 = \{c\}, X_3 = \{d\}$. The graph $G$ obtained is the same as in the proof of Th. 8, Fig. 12, except that concept nodes [x] corresponding to variables in $X_1$ are not linked to the nodes [xt] and [xf] representing their possible values.

First check that in this initial world, no constraint is violated, but the goal $Q$ cannot be satisfied. By applying once the evolution rule $E$, we try some valuation of the variables in $X_1$ and obtain a world $G_1$, that contains an answer to $Q$. But this world has to satisfy the positive constraint $C_+$, expressing that "for every valuation of the variables in $X_1 \cup X_2$, there must exist a valuation of the variables in $X_3$ such that the formula evaluates to *true*". If $G_1$ satisfies this constraint, it means that we have found (by applying $E$) a valuation of the variables in $X_1$ such that for all valuations of variables in $X_1 \cup X_2$ (which can be simplified in "for all valuations of variables in $X_2$", since there is only one such valuation for $X_1$), there is a valuation of the variables in $X_3$ such that the formula evaluates to *true*. Then there is an answer *yes* to the 3-$SAT_3$ problem. Conversely, suppose an answer *no* to the $\mathcal{SEC}$ problem. It means that for every world $G_1$ that can be obtained by applying the rule $E$, the constraint $C_+$ is violated (otherwise $Q$ could be projected into $G_1$ and the answer would be *yes*). Thus there is no assignment of the variables in $X_1$ satisfying the constraint, *i.e.* the answer to the 3-$SAT_3$ problem is *no*.

$\Sigma_3^P$-*completeness of* $\mathcal{SREC}$-*deduction.* $\mathcal{SREC}$-deduction stays in the same class of complexity as $\mathcal{SEC}$-deduction. Indeed, the question is "are there an $\mathcal{RE}$-derivation from $\mathcal{G}$ to $G'$ and a projection from $Q$ to a SG $G'$, such that for all $G_i$ of this derivation either equal to $\mathcal{G}$ or obtained by an immediate $\mathcal{E}$-derivation, for all $G_i'$ of this derivation derived from $G_i$ by an $\mathcal{R}$-derivation, for all constraint $C_j$, for all projection $\pi$ from $C_{j(0)}$ to $G_i'$, there exists an $\mathcal{R}$-derivation from $G_i'$ to a SG $G_i''$ and a projection $\pi'$ from $C_j$ to $G_i''$ s.t. $\pi'[C_{j(0)}] = \pi$?" and the lengths of all derivations are polynomial in the size of the input. When $\mathcal{R} = \emptyset$, one obtains $\mathcal{SEC}$-deduction, thus the $\Sigma_3^P$ completeness. □

Let us point out that, whereas in general case, deduction is more difficult in $\mathcal{SRC}$ (truly undecidable) than in $\mathcal{SEC}$ (semi-decidable), the converse holds for the particular case of range-restricted rules.

### 7.3 Particular Constraints

One may consider the case where not only rules but also constraints are restricted. Let us first consider the meaningful category of negative constraints.

**Theorem 13 (Complexity with negative constraints)** *Without any assumption on the rules in $\mathcal{E}$ or $\mathcal{R}$, but using only negative constraints:*

- $\mathcal{SGC}$-consistency *becomes co-NP-complete.*

- $\mathcal{SGC}$-deduction *becomes DP-complete.*

- $\mathcal{SRC}$-inconsistency *($\mathcal{SRC}$-consistency co-problem) becomes semi-decidable.*

- $\mathcal{SEC}$-deduction *remains semi-decidable.*

- $\mathcal{SRC}$-deduction *and* $\mathcal{SREC}$-deduction *remain truly undecidable.*

*Proof:*

*Co-NP-completeness of $\mathcal{SGC}$-consistency:* from NP-completeness of projection checking (th. 8).

*DP-completeness of $\mathcal{SGC}$-deduction:* this problem can be expressed as "is it true that $Q$ can be projected into $\mathcal{G}$ and that no constraint of $\mathcal{C}$ can be projected into $\mathcal{G}$?" thus belongs to DP. Now let us consider that $\mathcal{C}$ contains only one constraint. A reduction from 3-SAT to Projection (see f.i. the proof of th. 2) provides a straightforward reduction from SAT/UNSAT to $\mathcal{SGC}$-deduction (see f.i. Papadimitriou, 1994), thus the DP-completeness.

*Semi-decidability of $\mathcal{SRC}$-inconsistency:* To prove the inconsistency of a KB, we must find some violation of a constraint that will never be restored. But no violation of a negative constraint can ever be restored (further rule applications can only add information, thus more possible projections, and cannot remove the culprit one). So we only have to prove that one constraint of $\mathcal{C}$ can be deduced from $(\mathcal{G}, \mathcal{R})$: it is a semi-decidable problem. *Undecidability of $\mathcal{SRC}$-deduction* follows: we must prove that $Q$ can be deduced from $(\mathcal{G}, \mathcal{R})$, but that no constraint of $\mathcal{C}$ can.

The arguments proving semi-decidability of deduction in $\mathcal{SEC}$ and undecidability of deduction in $\mathcal{SREC}$ are the same as the ones used in the proof of Th. 10. □

The restriction to negative constraints decreases complexity of problems in the $\mathcal{SGC}$ model, but it does not help much as soon as rules are involved, since these problems remain undecidable. Combining range restricted rules and negative constraints, we obtain more interesting complexity results:

**Theorem 14 (Complexity with r.r. rules and negative constraints)** *If only range-restricted rules and negative constraints are present in the knowledge base:*

- $\mathcal{SRC}$-consistency *becomes co-NP-complete.*

- $\mathcal{SRC}$-deduction *becomes DP-complete.*

- $\mathcal{SEC}$-deduction *and* $\mathcal{SREC}$-deduction *become $\Sigma_2^P$-complete.*

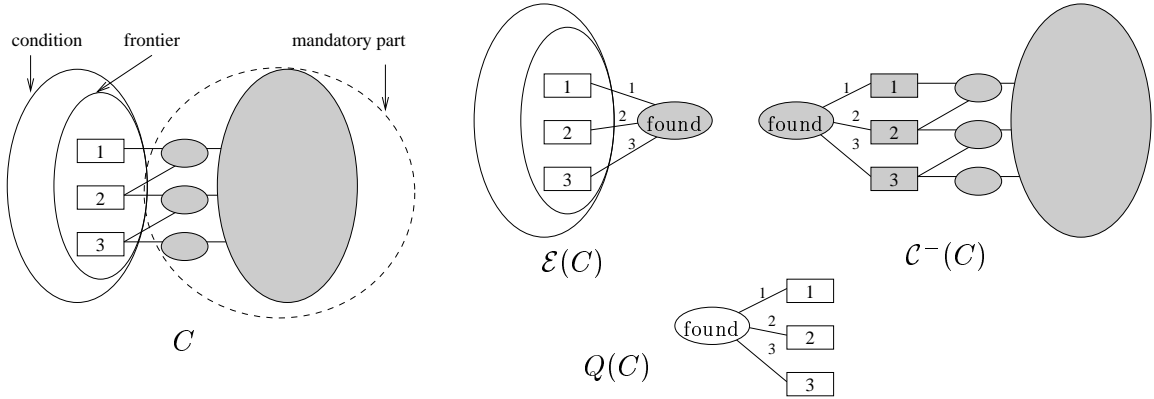

Figure 18: Transformation from $\mathcal{SGC}$-CONSISTENCY to a restricted $\mathcal{SEC}$-NON-DEDUCTION

*Proof:* Inconsistency in $\mathcal{SRC}$ admits a polynomial certificate, a derivation (of polynomial length) from $\mathcal{G}$ leading to a graph into which a constraint of $\mathcal{C}$ can be projected, and this projection. Inconsistency is thus in NP, and completeness follows from the particular case when $\mathcal{R}$ is empty. For deduction, we must prove that no constraint can be deduced from $(\mathcal{G}, \mathcal{R})$, but that $Q$ can. So the problem is in DP. For completeness, remark that the problem is still complete when $\mathcal{R}$ is empty (Th. 13).

To prove that $\mathcal{SEC}$-DEDUCTION with r.r. rules and negative constraints is $\Sigma_3^P$-complete, we will first show that it belongs to $\Sigma_2^P$, then exhibit a reduction from a $\Pi_2^P$-complete problem to its co-problem $\mathcal{SEC}$-NON-DEDUCTION (since co-$\Sigma_i^P = \Pi_i^P$).

$\mathcal{SEC}$-DEDUCTION corresponds to the language $L = \{x \mid \exists y_1 \forall y_2 \ R(x, y_1, y_2)\}$, where $x$ encodes an instance $(Q; (\mathcal{G}, \mathcal{E}, \mathcal{C}))$ of the problem, and $(x, y_1, y_2) \in R$ if $y_1$ encodes an $\mathcal{E}$-derivation from $\mathcal{G}$ to $G'$ and a projection from $Q$ to $G'$, and $y_2$ encodes a mapping from some constraint of $\mathcal{C}$ to $G'$ that is not a projection (note that if $G'$ does not violate any constraint, then no graph in the derivation from $\mathcal{G}$ to $G'$ does).

We exhibit now a reduction from the general $\mathcal{SGC}$-CONSISTENCY problem to $\mathcal{SEC}$-NON-DEDUCTION with r.r. rules and negative constraints. Let $(\mathcal{G}, \mathcal{C} = \{C\})$ be an instance of $\mathcal{SGC}$-CONSISTENCY (w.l.o.g., we restrict the problem to consider only one positive constraint). The transformation we consider builds an instance of $\mathcal{SEC}$-NON-DEDUCTION $(Q(C); (\mathcal{G}, \mathcal{E}(C), \mathcal{C}^-(C)))$ as follows. We call the *frontier* of the positive constraint $C$ the set of nodes in the trigger (*i.e.* colored by 0) having at least one neighbor in the obligation. The definition of colored graphs implies that frontier nodes are concept nodes (their neighbors are thus relation nodes). Let us denote these frontier nodes by $1, \ldots, k$. The evolution rule $\mathcal{E}(C)$ has for hypothesis the trigger of $C$, and for conclusion a relation node typed found, where found is a new $k$-ary relation type incomparable with all other types. The $i^{th}$ neighbor of this node is the concept node $i$. Check that $\mathcal{E}(C)$ is a range restricted rule. The negative constraint $\mathcal{C}^-(C)$ is the subgraph of $C$ composed of its obligation ($C_{(1)}$) added with nodes of the frontier and the relation node typed found, linked to the frontier nodes in the same way as above. Finally, the SG $Q(C)$ is made of one relation node typed found and its neighbors frontier nodes. This transformation is illustrated in Fig. 18.

W.l.o.g. we can assume that $\mathcal{G}$ is irredundant: in that case, $\mathcal{SGC}$-consistency is still $\Pi_2^P$-complete (see that the transformation used in the proof of Th. 8 produces an irredundant graph $G$). Now suppose that $(\mathcal{G}, C)$ is consistent: it means that either the trigger of $C$ does not project into $\mathcal{G}$, and in that case, the rule $\mathcal{E}(C)$ will never produce the needed (found) node, or every (existing) projection of the condition of $C$ into $\mathcal{G} = irr(\mathcal{G})$ can be extended to a projection of $C$ as a whole. So every application of $\mathcal{E}(C)$ produces a violation of $\mathcal{C}^-(C)$. In both cases $Q(C)$ cannot be deduced from the knowledge base. Conversely, suppose that $\mathcal{G}$ $\pi$-violates $C$, then the application of $\mathcal{E}(C)$ following $\pi$ produces a graph that does not violate $\mathcal{C}^-(C)$, and we can deduce $Q(C)$. □

The above theorem shows a decrease in complexity when general positive constraints are restricted to negative ones. $\mathcal{SGC}$-consistency falls from $\Pi_2^P$ to co-NP and, when also considering range restricted rules, $\mathcal{SRC}$-consistency falls from $\Pi_2^P$ to DP, and $\mathcal{SEC}$-deduction falls from $\Sigma_3^P$ to $\Sigma_2^P$. It would be interesting to exhibit particular cases of constraints, more general than negative ones, that make this complexity fall into intermediary classes (by example DP and $\Delta_2^P$ for $\mathcal{SGC}$-consistency). Some syntactic restrictions we defined for rules are good candidates: though a finite expansion set of constraints has no sense, let us consider *range restricted constraints*. Let us also define *disconnected constraints* as constraints where the trigger and the obligation are not connected; such constraints include the "topological constraints" used in (Mineau & Missaoui, 1997).

The following property highlights the relationships of these particular cases with negative constraints:

**Property 13** *Negative constraints are a particular case of both range-restricted constraints and disconnected constraints.*

*Proof:* As noticed in section 5, a negative constraint is equivalent to a positive constraint whose obligation is composed of one concept node of type NotThere, where NotThere is incomparable with all other types and does not appear in any SG except in $\mathcal{C}$ (it is thus a disconnected constraint). W.l.o.g. this node can be labeled by an individual marker (which, as NotThere, appears only in $\mathcal{C}$), thus leading to a constraint which is both disconnected and range-restricted. □

**Theorem 15 (Complexity with disconnected constraints)** *When $\mathcal{C}$ contains only disconnected constraints:*

- $\mathcal{SGC}$-consistency *becomes co-DP-complete.*

- $\mathcal{SRC}$-consistency *and* $\mathcal{SRC}$-deduction *remain undecidable, but* $\mathcal{SRC}$-consistency *becomes co-DP-complete when rules are range-restricted.*

- $\mathcal{SEC}$-deduction *remains semi-decidable, but becomes* $\Sigma_2^P$*-complete when rules are range-restricted.*

- $\mathcal{SREC}$-deduction *remains undecidable, but becomes* $\Sigma_2^P$*-complete when rules are range-restricted.*

*Proof:* $\mathcal{SGC}$-inconsistency belongs to DP, since we must prove that for one constraint there is a projection of its trigger and no projection of its obligation. Completeness is proved with a reduction from SAT/UNSAT (as in proof of Th. 13). $\mathcal{SGC}$-consistency is thus co-DP-complete.

Arguments for undecidability of $\mathcal{SRC}$-consistency, $\mathcal{SRC}$-deduction and $\mathcal{SREC}$-deduction, as well as semi-decidability of $\mathcal{SEC}$-deduction, are the same as in the proof of Th. 10: the constraints we used were already disconnected.

When rules are range-restricted, $\mathcal{SRC}$-inconsistency belongs to DP: we must prove that the trigger of the constraint can be deduced from $(\mathcal{G}, \mathcal{R})$, but not its obligation, and these problems belong respectively to NP and co-NP. Completeness comes from the particular case where $\mathcal{R}$ is empty. $\mathcal{SRC}$-consistency is thus co-DP-complete.

$\mathcal{SEC}$-deduction belongs to $\Sigma_2^P$ when rules involved are range-restricted. Though this property does not appear with an immediate formulation of the problem, it becomes obvious when the problem is stated as follows: "does there exist a sequence of graphs $\mathcal{G} = G_0, \ldots, G_p, G_{p+1}$, where $\mathcal{G} = G_0, \ldots, G_p$ is an $\mathcal{E}$-derivation and $G_{p+1}$ is the disjoint union of $G_p$ and $C_{(1)}$, a projection from $Q$ to $G_p$ and a projection from $C_{(1)}$ to a SG $G_k$, $0 \leq k \leq p+1$ such that for every graph $G_i, 0 \leq i < k$, $(*)$ for every mapping $\pi$ of $C_{(0)}$ into $G_i$, $\pi$ is not a projection ?". Notice that no $G_i$ before $G_k$ in such a sequence triggers the constraint ($C_{(0)}$ does not project into $G_i$) and that all $G_i, i \geq k$, satisfy it (since $C_{(1)}$ projects to $G_i$), thus all $G_i$ of the sequence are consistent. $G_{p+1}$ ensures that $C_{(1)}$ projects into at least one graph of the sequence, which allows the above formulation of the problem. Completeness follows from the particular case of negative constraints.

Proof for $\mathcal{SREC}$-deduction in the case of range restricted rules is similar: in the expression of the problem above, the derivation is now an $(\mathcal{E} \cup \mathcal{R})$-derivation, the $G_i$ considered are only the ones obtained after the application of a rule from $\mathcal{E}$, and $(*)$ "for every mapping $\pi$ of $C_{(0)}$ into $G_i$" is replaced by "for every graph that can be $\mathcal{R}$-derived from $G_i$". □

Unfortunately, range-restricted constraints are trickier to study: intuitively, consistency checking should become easier than with general constraints, but the role of irredundancy is still unclear. Though it is easy to check that $\mathcal{SGC}$-deduction with range restricted constraints is at least DP-hard (transformation from SAT/UNSAT) and we have proven (though it is not included in this paper) that it is in $\Delta_2^P$ (i.e. $P^{NP}$), we did not manage to achieve an exact complexity result for this problem.

We did not either manage to assign a complexity class for the $\mathcal{SGC}^d$-deduction and $\mathcal{SR}^{rr}\mathcal{C}^d$-deduction problems, though both problems trivially belong to $\Delta_2 P$.

Complexity results obtained in this paper are summarized in table. 1. We also present in Fig. 19 a "complexity map" emphasizing the relationships between problems. In this figure, if $E$ denotes a set of bicolored graphs (rules or constraints), $E^{fes}$, $E^{rr}$, $E^d$ respectively denote its restriction to a finite expansion set, range restricted elements, or disconnected elements. $\mathcal{C}^-$ denotes a set of negative constraints. All problems represented are complete for their class. Edges are directed from bottom to top. An edge from a problem P1 to a problem P2 means that P1 is a particular case of P2. Moreover, in order that the map remains readable, problems which are intermediate between two problems P1 and P2 of the same complexity class, do not appear in the figure. The complexity of such problems can be obtained by "classifying" them in the hierarchy. For instance, $\mathcal{SEC}$-deduction is more general than $\mathcal{SR}$-deduction (which is obtained if $\mathcal{C} = \emptyset$) and more specific than $\mathcal{SR}^{fes}\mathcal{EC}$-

| | General case | $\mathcal{R}$ fes | $\mathcal{E}$ fes | $\mathcal{R} \cup \mathcal{E}$ fes | $\mathcal{R}\&\mathcal{E}$ r.r. | $\mathcal{C}^-$ | $\mathcal{R}\&\mathcal{E}$ r.r., $\mathcal{C}^-$ | $\mathcal{R}\&\mathcal{E}$ r.r., $\mathcal{C}^d$ |
|---|---|---|---|---|---|---|---|---|
| $\mathcal{SG}$-ded. | NP-C | / | / | / | / | | / | / |
| $\mathcal{SGC}$-cons. | $\Pi_2^P$-C | / | / | / | | co-NP-C | co-NP-C | co-DP-C |
| $\mathcal{SGC}$-ded. | $\Pi_2^P$-C | / | | / | | DP-C | DP-C | ? |
| $\mathcal{SR}$-ded. | semi-dec. | dec. | | dec. | NP-C | | NP-C | NP-C |
| $\mathcal{SRC}$-cons. | undec. | dec. | / | dec. | $\Pi_2^P$-C | co-semi-dec. | co-NP-C | co-DP-C |
| $\mathcal{SRC}$-ded. | undec. | dec. | | dec. | $\Pi_2^P$-C | undec. | DP-C | ? |
| $\mathcal{SEC}$-ded. | semi-dec. | / | dec. | dec. | $\Sigma_3^P$-C | semi-dec. | $\Sigma_2^P$-C | $\Sigma_2^P$-C |
| $\mathcal{SREC}$-ded. | undec. | semi-dec. | undec. | dec. | $\Sigma_3^P$-C | undec. | $\Sigma_2^P$-C | $\Sigma_2^P$-C |

Table 1: Summary of Complexity Results

DEDUCTION (which adds the set $R^{fes}$), and, since these problems are both semi-decidable, so is $\mathcal{SEC}$-deduction.

## 8. Related Works

One interesting relationship from an algorithmic viewpoint is with the CSP framework. Recall the input of a constraint satisfaction problem (CSP) is a constraint network, composed of a set of variables, sets of possible values for the variables (called their domains) and a set of constraints between the variables. The question is whether there is a solution to the CSP, i.e. an assignment of values to the variables that satisfies the constraints.

The constraints involved in a classical CSP are simpler than ours. Actually, CSP corresponds to the $\mathcal{SG}$-DEDUCTION (projection) problem.

Several authors noticed the strong equivalence between CSP and LABELLED GRAPH HOMOMORPHISM (Given two labeled graphs $G$ and $H$, is there a homomorphism from $G$ to $H$?). As far as we know, the first paper on this subject was (Feder & Vardi, 1993). In (Mugnier, 2000) correspondences are detailed from PROJECTION (Given two SGs $G$ and $H$ is there a projection from $G$ to $H$ ?) to CSP, and reciprocally (developing the ones presented in (Mugnier & Chein, 1996)). Let us outline the ideas of the transformation from CSP to PROJECTION, called $C2P$. Consider a constraint network $P$. $P$ is transformed into two SGs $G$ and $H$ as follows. $G$ translates the *structure* of $P$: each concept node is generic and corresponds to a variable and each relation node corresponds to a constraint (its $i$th neighbor is the concept node corresponding to the $i$th variable of the constraint). $H$ represents the *constraint definitions*: there is one individual concept node for each value of a variable domain, and one relation node for each tuple of compatible values. Roughly said, there is a solution to $P$ if there is a mapping from variables (concept nodes of $G$) to values (concept nodes of $H$) that satisfies the constraints (maps relation nodes of $G$ onto relation nodes of $H$), *i.e.* a projection from $G$ to $H$. The same result has been achieved independently in the Attributed Graph Grammar formalism by Rudolf (1998).

One could also see CSP as a particular case of $\mathcal{SGC}$-CONSISTENCY: indeed, there is a projection from a SG $G$ into a SG $H$ if and only if $H$ satisfies the positive constraint with an empty trigger and $G$ as its obligation.

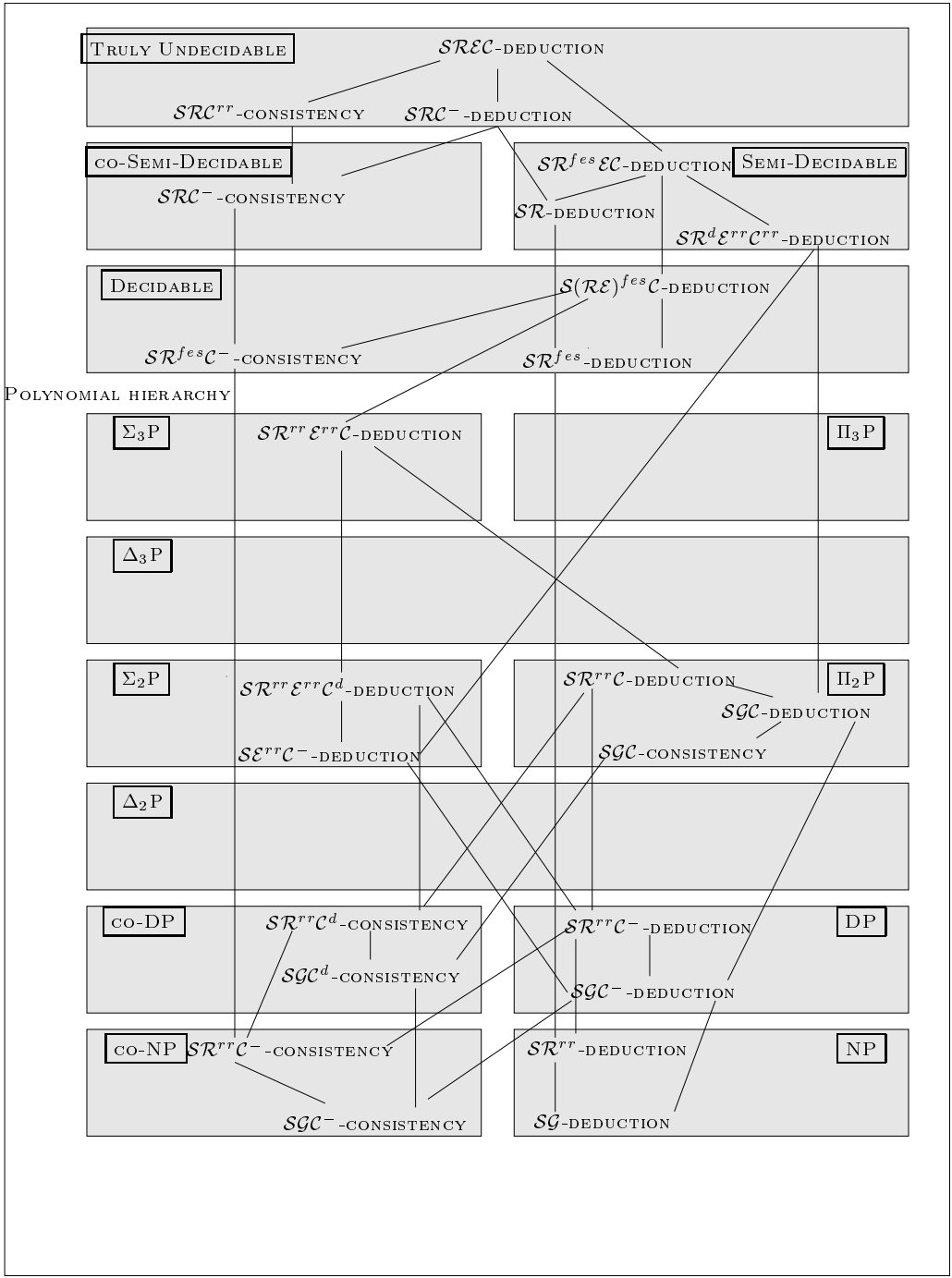

Figure 19: Complexity Results: a Geography

Now, in order to deal with incomplete knowledge Fargier et al. (1996) [1] extend the CSP framework to mixed-CSP. In a mixed-CSP the set of variables is decomposed into controllable and uncontrollable variables, say $X$ and $\Delta$. The MIXED-SAT problem asks whether a binary mixed-CSP is consistent, which can be reformulated as follows: is it true that every solution to the subnetwork induced by $\Delta$ can be extended to a solution of the whole network? MIXED-SAT is shown to be $\Pi_2^P$-complete. This result provides us another proof of $\Pi_2^P$-completeness for $\mathcal{SGC}$-consistency. Indeed, any mixed-CSP can be translated into an instance of $\mathcal{SGC}$-CONSISTENCY. Using the $C2P$ reduction described above, the mixed-CSP is mapped to SGs $G$ and $H$. $G$ is then provided with two colors, giving a positive constraint $C$, whose trigger is the subgraph corresponding to the subnetwork induced by $\Delta$. The mixed-CSP is consistent if and only if $H$ satisfies $C$. Fig. 20 illustrates this transformation. The constraint network is composed of the two variable sets $X = \{x_1, x_2\}$ and $\Delta = \{l_1, l_2\}$ and three constraints $C_1$, $C_2$ and $C_3$. $x_1$ and $x_2$ have same domain $\{1, 2, 3\}$ and $l_1$ and $l_2$ have same domain $\{a, b\}$. The constraint definitions are given in the figure. All concept types are supposed to be incomparable.

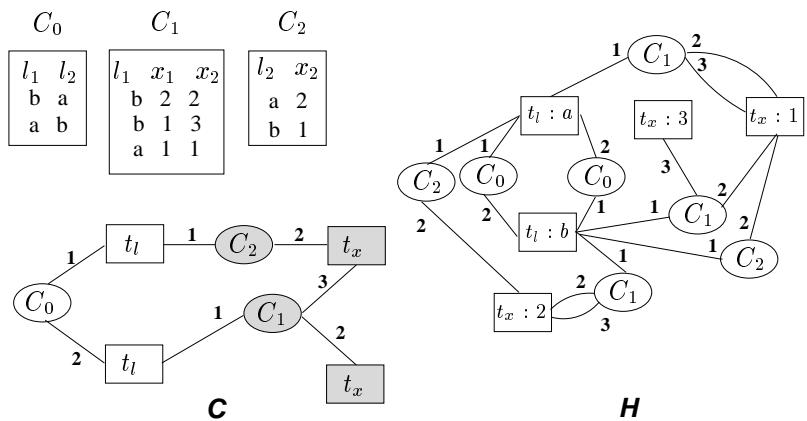

Figure 20: Transformation from MIXED-SAT to $\mathcal{SGC}$-CONSISTENCY

Let us relate our definitions to other definitions of constraints found in the CG literature. Our constraints (let us call them SG-constraints) are a particular case of the minimal descriptive constraints defined in (Dibie et al., 1998): a minimal descriptive constraint can be seen as a set of SG-constraints with the same trigger; its intuitive semantics is "if $A$ holds so must $B_1$ or $B_2$ or ... $B_k$". A SG satisfies a minimal descriptive constraint if it satisfies at least one element of the set. Note that the "disjunction" does not increase the complexity of the consistency check relative to SG-constraints. The proof of theorem 8 (complexity of $\mathcal{SGC}$-CONSISTENCY) can be used to show that consistency of minimal descriptive constraints is also $\Pi_2^P$-complete. Dibie et al. (1998) have pointed out that minimal descriptive constraints generalize most constraints found in the CG literature. Actually, these latter constraints are also particular cases of SG-constraints (for instance, as already noticed, the topological constraints used by Mineau & Missaoui, 1997 are disconnected SG-constraints). Let us add that, in these CG works, constraints are used to check consistency of SGs solely and not of richer knowledge bases composed of rules (as in $\mathcal{SRC}$) and they are not integrated into more complex reasonings (as in $\mathcal{SEC}$ or in $\mathcal{SREC}$).

---

1. We thank Christian Bessière for this reference.

There should be other connections with works about verification of knowledge bases composed of logical rules (for instance Horn rules), namely with the works of Levy and Rousset (1996), in which constraints are TGDs, thus have the same form than ours, but we did not find direct relationships between their framework and ours.

As both models are rooted in semantic networks, comparing conceptual graphs and descriptions logics is a problem that has often been issued. Baader, Molitor, and Tobies (1999) have identified a fragment of the $\mathcal{SG}$ model (where simple graphs are restricted to those having a tree-like structure, but conjunctive types are allowed) with a language called $\mathcal{ELIRO}$[1]: this equivalence has led to a new tractability result in description logics. However, trying to identify larger fragments seems to be a dead-end: as pointed out by Mugnier (2000), projection cannot handle negation on primitive types. On the other hand, even the most expressive description logics cannot express the whole FOL($\wedge, \exists$) fragment (Borgida, 1996). Encoding some existing description logics into models of the $\mathcal{SG}$ family is an interesting perspective, that could allow one to identify new decidable classes for our models, add type expressiveness to conceptual graphs, and may be cycles in the description of DLs concepts.

## 9. Conclusion

We have proposed a family of models that can be seen as the basis of a generic modeling framework. Main features of this framework are the following: a clear distinction between different kinds of knowledge, that fit well with intuitive categories, a uniform graph-based language that keeps essential properties of the SG model, namely readability of objects as well as reasonings. We guess this latter point is particularly important for the usability of any knowledge based system. In our framework, all kinds of knowledge are graphs easily interpreted, and reasonings can be graphically represented in a natural manner using the graphs themselves, thus explained to the user on its own modelization.

Technical contributions, w.r.t. previous works on conceptual graphs, can be summarized as follows:

- the representation of different kinds of knowledge as colored SGs: facts, inference rules, evolution rules and constraints.

- the integration of constraints into a reasoning model; more or less similar notions of a constraint had already been introduced but were only used to check consistency of a simple graph (as in the $\mathcal{SGC}$ model). The complexity of consistency checking was not known.

- a systematic study of the obtained family of models with a complexity classification of associated consistency/deduction problems, including the study of particular cases of rules and constraints, which provide interesting complexity results.

We also established links between consistency checking and FOL deduction, translating the consistency/deduction problems in terms of FOL deduction. It should be noticed that the operational semantics of models combining rules and constraints, namely $\mathcal{SREC}$, $\mathcal{SRC}$ and $\mathcal{SEC}$, is easy to understand but we were not able to give a global logical semantics. Indeed, there is an underlying non monotonic mechanism whose logical interpretation should require non standard logics. The definition of a logical semantics for these models is thus an open problem.

## Acknowledgments.

We are indebted to Michel Chein and Geneviève Simonet for their careful reading of an earlier version of this work and helpful comments. We also wish to thank Marie-Christine Rousset and Pierre Marquis, as well as anonymous referees, for their interesting suggestions and some corrections.